\documentclass[final]{IEEEtran}

\usepackage{graphicx}
\usepackage[cmex10]{amsmath}
\usepackage{algorithm2e}
\usepackage{array}
\usepackage[tight,footnotesize]{subfigure}
\usepackage{url}
\usepackage{framed, color}
\usepackage{tikz}
\usetikzlibrary{backgrounds,fit,shapes.multipart,plotmarks,patterns,shapes,arrows}

\usepackage[noadjust]{cite}

\definecolor{pu1}{rgb}{0.25,0.25,0.25}
\definecolor{pu2}{rgb}{0, 1, 0}
\definecolor{pu3}{rgb}{0,1,1}
\definecolor{pu4}{rgb}{0,0.5,0}
\definecolor{pu5}{rgb}{1,0,1}
\definecolor{pu6}{rgb}{0.65,0.33,0.15}
\definecolor{pu7}{rgb}{0.5,0,0.5}
\definecolor{pu8}{rgb}{1,0,0}
\definecolor{pu9}{rgb}{1,1,0}

\definecolor{pc1}{rgb}{0, 0, 1}
\definecolor{pc2}{rgb}{0, 0.5, 0}
\definecolor{pc3}{rgb}{0, 1,0}
\definecolor{pc4}{rgb}{1, 0, 0}
\definecolor{pc5}{rgb}{0.65,0.33,0.15}
\definecolor{pc6}{rgb}{0.25, 0.25, 0.25}
\definecolor{pc7}{rgb}{0.5, 0, 0.5}
\definecolor{pc8}{rgb}{1, 0.45, 0}
\definecolor{pc9}{rgb}{1, 1, 0}

\definecolor{i1}{rgb}{1,0,0}
\definecolor{i2}{rgb}{0,0.5,0.5}
\definecolor{i3}{rgb}{0,1,1}
\definecolor{i4}{rgb}{0,0,0.5}
\definecolor{i5}{rgb}{0,1,0}
\definecolor{i6}{rgb}{0,0.5,0}
\definecolor{i7}{rgb}{0.5,0.5,0}
\definecolor{i8}{rgb}{1,1,0}
\definecolor{i9}{rgb}{0,0,1}
\definecolor{i10}{rgb}{0.4,0.08,0.4}
\definecolor{i11}{rgb}{0.75, 0, 0.75}
\definecolor{i12}{rgb}{1,0,1}
\definecolor{i13}{rgb}{0.55,0.3, 0}
\definecolor{i14}{rgb}{0.5,0,0}
\definecolor{i15}{rgb}{0.5,0.5,0.5}
\definecolor{i16}{rgb}{0,0,0}

\definecolor{be_green}{rgb}{0.549019607843137, 0.705882352941177, 0.0392156862745098} 
\definecolor{be_blue}{rgb}{0.235294117647059, 0.705882352941177, 0.862745098039216} 
\definecolor{blue2}{rgb}{0.2, 0.55, 0.59} 

\newcommand{\ie}{i.e.}

\newcommand{\eg}{e.g.}

\newcommand{\name}{shape-DC}

\renewcommand{\(}{\left(}
\renewcommand{\)}{\right)}

\renewcommand{\vector}[1]{{\mbox{\boldmath$#1$}}}

\newcommand{\m}[1]{{\mbox{{\fontencoding{T1}\sffamily\slshape{#1\/}}}}}

\newcommand{\mR}{{\rm I\!R}}

\newcommand{\ical}[1]{{\mbox{\usefont{OT1}{pzc}{m}{it}{#1}}}}

\makeatletter
\newcommand{\removelatexerror}{\let\@latex@error\@gobble}
\makeatother

\begin{document}

\title{Shapelet-based Sparse Representation for Landcover Classification of Hyperspectral Images}

\author{Ribana Roscher,~\IEEEmembership{Member,~IEEE,} 
        Bj\"orn Waske,~\IEEEmembership{Member,~IEEE}
\IEEEcompsocitemizethanks{\IEEEcompsocthanksitem 
The authors are with the Institute of Geographical Sciences, Remote Sensing and Geoinformatics, Freie Universit\"at Berlin, Malteserstr. 74--100, 12249 Berlin, Germany. \protect\\
E-mail: ribana.roscher@fu-berlin.de, phone: (+49 30) 838-70350
}
}

\markboth{IEEE Transactions on Geoscience and Remote Sensing,~Vol.~54, No.~3, March~2016}%
{Roscher \MakeLowercase{\textit{et al.}}: Shapelet-Based Sparse Representation for Landcover Classification of Hyperspectral Images}

\maketitle

\begin{abstract}
This paper presents a sparse representation-based classification approach with a novel dictionary construction procedure.
By using the constructed dictionary sophisticated prior knowledge about the spatial nature of the image can be integrated.
The approach is based on the assumption that each image patch can be factorized into characteristic spatial patterns, also called shapelets, and patch-specific spectral information.
A set of shapelets is learned in an unsupervised way and spectral information are embodied by training samples. 
A combination of shapelets and spectral information are represented in an undercomplete spatial-spectral dictionary for each individual patch, where the elements of the dictionary are linearly combined to a sparse representation of the patch.
The patch-based classification is obtained by means of the representation error.
Experiments are conducted on three well-known hyperspectral image datasets.
They illustrate that our proposed approach shows superior results in comparison to sparse representation-based classifiers that use only limited spatial information and behaves competitively with or better than state-of-the-art classifiers utilizing spatial information and kernelized sparse representation-based classifiers. 
\end{abstract}

\begin{IEEEkeywords}
sparse coding, sparse representation, dictionary construction, shapelets, hyperspectral
\end{IEEEkeywords}

\IEEEpeerreviewmaketitle


\section{Introduction}
Land cover classification is one of the most common task for remote
sensing applications and the development of adequate classification strategies is an ongoing research field. 
In this context, hyperspectral imagery is probably the most valuable as well as challenging single data source. 
Hyperspectral sensors provide detailed and spectrally continuous spatial information, enabling the discrimination between spectrally similar land cover classes (\eg, \cite{Camps-Valls2014, Bioucas-Dias2013}). 
However, it is well known that data dimensionality and high redundancy between individual spectral bands cause challenges during data analysis, for example, the performance of standard supervised classifiers is often limited in terms of classification accuracy. 
Therefore alternative methods such as support vector machines (SVM, \eg, \cite{Mountrakis2011, Melgani2004}), ensemble-based learning (\eg, \cite{Waske2010, Waske2009}) or classifiers based on multinomial logistic regression (\eg, \cite{Roscher2012, Li2010, Demir2007}) have been successfully used for hyperspectral image classification.

Another successful development for classification of remote sensing data is the integration of spatial/contextual information (\eg, \cite{Camps-Valls2014, Schindler2012, Fauvel2013}).
In this way, the spatial correlation between adjacent pixels can be taken into account.
For instance, Camps-Valls et al. \cite{Camps-Valls2006} and Tuia et al. \cite{Tuia2010} introduced an SVM with additional spatial information by means of a composite kernel.
This kernel is a combination of a spectral kernel derived from features extracted from the pixel itself and a contextual kernel comprising features from the surrounding area such as the mean or the standard deviation.
An alternative approach was proposed in \cite{Benediktsson2005, Fauvel2008,DallaMura2010}, which uses mathematical morphology in order to represent the spatial relationship between pixels.
Both approaches are combined by Li et al. \cite{Li2013} to generalized composite kernels for multinomial logistic regression.
Besides the aforementioned pixel-based classification strategies, classification can alternatively be performed on pre-defined regions obtained by image segmentation (\eg, \cite{Blaschke2010,Bruzzone2006,Kettig1976}).
Another common method to incorporate both spectral and spatial information is the usage of a Markov random field, which utilizes an additional spatial term in order to favour class smoothness for the final classification map (\eg, \cite{Li2014,Schindler2012,Tarabalka2010}).

Besides the aforementioned classifiers, sparse representation-based classifiers have been recently introduced in the context of hyperspectral image classification, showing state-of-the-art classification performance.
Sparse representation-based classifiers should not be confused with sparse classifiers such as SVM, since their underlying concept is different.
A sparse representation-based classifier assumes that each pixel can be reconstructed by a sparsely weighted linear combination of a few basis vectors taken from a so-called dictionary.
The dictionary is constructed from a set of representative samples, for instance the training data, and is either directly embodied by these samples (\eg, \cite{Soltani-Farani2013, Chen2011}) or learned from them (\eg, \cite{Yang2014,Charles2011,Castrodad2011}).
In the context of supervised classification each dictionary element also provides a class label, which is used for the classification of the sparsely represented sample of interest. 

For sparse representation and classification of RGB- and gray-valued image data, generally, the dictionary elements are representative vectorized image patches, which are derived from the labeled training data (\eg, \cite{Mairal2012, Jenatton2011}).
In this way, spectral as well as spatial information are integrated in the dictionary.
Commonly, these dictionaries are chosen to be overcomplete, \ie, the number of dictionary elements is larger than the dimension of elements, in order to have a high coding efficiency and a high approximation ability (\cite{Mairal2012,Kreutz-Delgado2003,Lewicki2000}).
Regarding to high dimensionality, overcomplete sparse representations for hyperspectral data are challenging due to the definition of representative dictionary elements, particularly, if patches are used rather than single pixels to integrate spatial information.
Apart from this, generally, there is a lack of labeled image patches if a labeled dictionary is used for classification.

Against this background, several sparse representation-based classifiers have been proposed which alternatively use pixel-based dictionaries and incorporate spectral and spatial information by using structured priors.
Various approaches assume that remote sensing images are smooth, \ie, neighboring pixels tend to have similar spectral characteristics, and thus, exploit the spatial correlation within the sparse coding procedure.
For example, Chen et al. \cite{Chen2013} successfully applied a joint sparsity model for hyperspectral image classification.
Here, neighboring pixels within an image patch are sparsely represented by a common set of dictionary elements.
This is similar to the multiple measurement vector problem, which is widely considered in the field of compressive sensing (\eg, \cite{Ziniel2013}, \cite{Davies2012}).
Chen et al. \cite{Chen2013} realize the determination of the sparse weights via the greedy optimization procedure simultaneous orthogonal matching pursuit (SOMP, \cite{Tropp2006}), which has also been used, for example, by Fu et al. \cite{Fu2014} and Aravind et al. \cite{Aravind2011} for hyperspectral image analysis.
This algorithm is an extension to orthogonal matching pursuit (OMP, \cite{Pati1993}), which is a standard method for solving the sparse coding task.
However, the mentioned approach is restrictive since it only assumes homogeneous regions with similar spectral information in each patch.
Therefore, actual class transitions (\ie, boundaries between classes within the image), and regions of the same class, which show different spectral properties, cannot be considered. 
In order to mitigate the influence of this problem, \eg, Yuan et al. \cite{Yuan2013} and Srinivas et al. \cite{Srinivas2013} introduced sparse representation-based classifiers utilizing different weights for all neighboring pixels depending on their similarity to the pixel of interest. 
Finally, Sun et al. \cite{Sun2014} presented several structured priors, which exploit the spatial dependencies between neighboring pixels, but also the inherent structure of the dictionary.

All the above mentioned approaches are based on the assumption that a pixel of interest can be approximated by a linear combination of representative training pixels.
That means, the dictionary contains the actual training data and most of the works treat the dictionary elements independently from each other without any prior spatial assumptions.
In a similar way, \eg, the approach in \cite{Iordache2014} and various works presented in \cite{Bioucas-Dias2012} use spectral libraries with pure spectra as dictionary, which is often used for spectral unmixing.
In contrast to the utilization of user-defined dictionaries, several authors consider a dictionary learning or dictionary construction task, \ie, finding the best dictionary to a dataset. 
As already stated in \cite{Arora2014} and \cite{Rubinstein2010}, the determination of the dictionary is challenging yet crucial for the success of the sparse representation.
For example, Charles et al. \cite{Charles2011} learn the dictionary elements by adapting randomly chosen pixels from the scene with regard to their sparse coding ability causing that the dictionary elements do not longer represent material spectra.
This shows that pure spectra do not necessarily possess the best coding ability.
The work was extended in \cite{Charles2014} by incorporating additional spatial regularity, so that the sparse parameter vector of a pixel is influenced by the sparse weights of the neighbors, but without forcing on spatial homogeneity.
In \cite{Soltani-Farani2013} the authors learn the dictionary elements using the same joint sparsity assumption as in \cite{Chen2011} and so-called contextual groups, \ie, non-overlapping image patches.
The estimated sparse parameter vectors are then utilized as features for a linear SVM classifier.

Considering the same task, this paper presents a sparse representation-based approach including a novel dictionary construction approach for hyperspectral image classification that explicitly introduces prior knowledge about the spatial nature of the image.
Both the spectral information and the spatial knowledge are combined to a patch-based dictionary containing representative vectorized image patches.
The patches are specifically constructed for the image patch which is to be classified.
It is more comprehensive than approaches that only assume a patch-based homogeneous neighborhood, since it also accounts for multiple homogeneous regions within a patch. 
The construction of the dictionary is based on a method which originally has been introduced for RGB images in \cite{Chua2012} and is adapted and extended to meet the challenges of hyperspectral image analysis.
It assumes that images generally contain repeated patterns, \ie, certain local arrangements of pixels with similar features.
If the most representative patterns are known (in form of patches in our case), the image can be approximated based on the factorization into these patterns and adequate spectral information, which are chosen to be the training data.
Shapelets and spectral information are combined to a patch-specific spatial-spectral dictionary.
Please note that unlike many dictionary learning approaches, the dictionary estimation and the reconstruction step for classification are decoupled. 
Additionally, unlike many approaches, the constructed dictionary turns out to be highly undercomplete, which means the number of dictionary elements is smaller than the dimension of the dictionary elements.
As already stated by Mairal et al. \cite{Mairal2012}, overcomplete dictionaries are often preferred for reconstructions tasks, but for classification tasks a perfect reconstruction is not necessarily required as long as the dictionary has enough discrimination power.
Also Moody et al. \cite{Moody2013} show that undercomplete dictionaries are suitable for a successful landcover classification.
Our combined dictionary is used for a sparse coding procedure aiming at the sparse representation of the image patches.
Hereby, it is assumed that each patch can be represented by a sparsely weighted linear combination of elements out of this patch-specific dictionary.
The estimated sparse weights and their assigned dictionary elements are then used for classification of the patch.

Overall, the objective of the paper is the presentation of a novel labeled patch-based dictionary construction approach exploiting spatial as well as spectral information used for a sparse representation-based classifier.
In order to evaluate the potential of such a concept, we focus on the following research questions: (i) Is there a difference in accuracy compared to support vector machines with composite kernel and sparse representation-based classifiers with spatial information; (ii) What is the impact of the hyperparameters, namely number of shapelets and patchsize, on the accuracy and stability of our proposed method in terms of the classification result? In order to answer these research questions, the specific objective of our study is the classification of three widely-used different hyperspectral datasets.

The following section describes the basic sparse coding procedure and the shapelet-based sparse representation scheme.
Moreover the patch-specific spatial-spectral dictionary construction is presented.
Sec. \ref{sec:experiments} and Sec.\ref{sec:results} demonstrate the experiments and discuss the results in order to show the performance of our proposed approach. 

\section{Methods}
\label{sec:methods}
Let $\m I$ be an $M$-band hyperspectral image containing $J$ overlapping patches $\vector X_j$ of size $\sqrt{Z} \times \sqrt{Z}\times M$ with $Z$ being the number of pixels in the patch.
The task is to assign a class label to each pixel by classifying all image patches and combine the results by a voting scheme.
Each patch is represented via a sparsely weighted linear combination of dictionary elements and the sparse weighting vector is determined using a sparse coding procedure (see Sec. \ref{sec:basic}).
A spatial-spectral dictionary is constructed utilizing a set of learned shapelets and a spectral sample set (see Sec. \ref{sec:dict}).
The construction strategy is based on the assumption that each image patch can be factorized into shape and spectral information (see Sec. \ref{sec:scheme}).

\subsection{Sparse Coding}
\label{sec:basic}
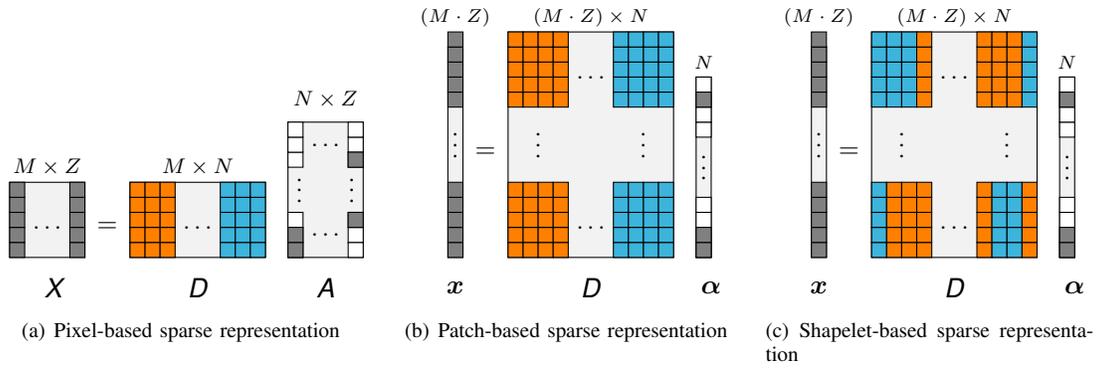
\begin{figure*}[ht]
\centering
\subfigure[Pixel-based sparse representation]{
\label{fig:SRa}
\begin{tikzpicture}    
\draw [fill=gray!10] (-1.6, 0) rectangle +(1.0, 1.0);	
\foreach \x in {0,...,4}
	\draw [fill=gray] (-1.6, \x/5) rectangle +(0.2, 0.2);
\foreach \x in {0,...,4}
	\draw [fill=gray] (-0.8, \x/5) rectangle +(0.2, 0.2);
	
\draw [fill=gray!10] (0, 0) rectangle +(1.8, 1.0);

\foreach \y in {0,...,2}
	\foreach \x in {0,...,4}
		\draw [fill=orange] (\y/5, \x/5) rectangle +(0.2, 0.2);		
\foreach \y in {0,...,2}
	\foreach \x in {0,...,4}
		\draw [fill=be_blue] (\y/5+1.2, \x/5) rectangle +(0.2, 0.2);	

\draw [fill=gray!10] (2.1, 0) rectangle +(1.0, 1.8);	
\foreach \x in {0,...,2}
	\draw [fill=white] (2.1, \x/5) rectangle +(0.2, 0.2);
\foreach \x in {0,...,2}
	\draw [fill=white] (2.9, \x/5) rectangle +(0.2, 0.2);
\foreach \x in {6,...,8}
	\draw [fill=white] (2.1, \x/5) rectangle +(0.2, 0.2);
\foreach \x in {6,...,8}
	\draw [fill=white] (2.9, \x/5) rectangle +(0.2, 0.2);
	
\draw [fill=gray] (2.1, 0.0) rectangle +(0.2, 0.2);	
\draw [fill=gray] (2.1, 0.2) rectangle +(0.2, 0.2);	
\draw [fill=gray] (2.9, 0.4) rectangle +(0.2, 0.2);			
\draw [fill=gray] (2.9, 1.2) rectangle +(0.2, 0.2);	

  \draw (-1.1, 1.2) node {\footnotesize $M\times Z$};
  \draw (0.9, 1.2) node {\footnotesize $M\times N$};
  \draw (2.6, 2.1) node {\footnotesize $N\times Z$};   
  \draw (-1.0, -.4) node {$\m X$};
  \draw (-0.3, 0.4) node {$=$};
  \draw (0.9, -.4) node {$\m D$};
  \draw (2.6, -.4) node {$\m A$};
  \draw (0.9, 0.4) node {\small $\boldmath\ldots$};
  \draw (-1.1, 0.4) node {\small $\boldmath\ldots$};
  \draw (2.6, 0.3) node {\small $\boldmath\ldots$};
  \draw (2.6, 1.5) node {\small $\boldmath\ldots$};
  \draw (2.25, 1) node {\small $\boldmath\vdots$};
  \draw (2.95, 1) node {\small $\boldmath\vdots$};
  
\end{tikzpicture}}
\quad
\subfigure[Patch-based sparse representation]{
\label{fig:SRb}
\begin{tikzpicture}     
\draw [fill=gray!10] (-1.6, 0) rectangle +(0.2, 2.8);
\foreach \x in {0,...,4}
	\draw [fill=gray] (-1.6, \x/5) rectangle +(0.2, 0.2);
\foreach \x in {0,...,4}
	\draw [fill=gray] (-1.6, \x/5+2.0) rectangle +(0.2, 0.2);
	
\draw [fill=gray!10] (-0.8, 0) rectangle +(2.2, 3.0);
\foreach \x in {0,...,4}
	\draw [fill=orange] (-0.8, \x/5) rectangle +(0.2, 0.2);	
\foreach \x in {0,...,4}
	\draw [fill=orange] (-0.8, \x/5+2.0) rectangle +(0.2, 0.2);	
\foreach \x in {0,...,4}
	\draw [fill=orange] (-0.6, \x/5) rectangle +(0.2, 0.2);	
\foreach \x in {0,...,4}
	\draw [fill=orange] (-0.6, \x/5+2.0) rectangle +(0.2, 0.2);	
\foreach \x in {0,...,4}
	\draw [fill=orange] (-0.4, \x/5) rectangle +(0.2, 0.2);	
\foreach \x in {0,...,4}
	\draw [fill=orange] (-0.4, \x/5+2.0) rectangle +(0.2, 0.2);	
\foreach \x in {0,...,4}
	\draw [fill=orange] (-0.2, \x/5) rectangle +(0.2, 0.2);	
\foreach \x in {0,...,4}
	\draw [fill=orange] (-0.2, \x/5+2.0) rectangle +(0.2, 0.2);	
\foreach \x in {0,...,4}
	\draw [fill=be_blue] (0.6, \x/5) rectangle +(0.2, 0.2);	
\foreach \x in {0,...,4}
	\draw [fill=be_blue] (0.6, \x/5+2.0) rectangle +(0.2, 0.2);
\foreach \x in {0,...,4}
	\draw [fill=be_blue] (0.8, \x/5) rectangle +(0.2, 0.2);	
\foreach \x in {0,...,4}
	\draw [fill=be_blue] (0.8, \x/5+2.0) rectangle +(0.2, 0.2);
\foreach \x in {0,...,4}
	\draw [fill=be_blue] (1, \x/5) rectangle +(0.2, 0.2);	
\foreach \x in {0,...,4}
	\draw [fill=be_blue] (1, \x/5+2.0) rectangle +(0.2, 0.2);
\foreach \x in {0,...,4}
	\draw [fill=be_blue] (1.2, \x/5) rectangle +(0.2, 0.2);	
\foreach \x in {0,...,4}
	\draw [fill=be_blue] (1.2, \x/5+2.0) rectangle +(0.2, 0.2);
							
\draw [fill=gray!10] (1.7, 0) rectangle +(0.2, 2.4);
\foreach \x in {0,...,3}
	\draw [fill=white] (1.7, \x/5) rectangle +(0.2, 0.2);
\foreach \x in {8,...,11}
	\draw [fill=white] (1.7, \x/5) rectangle +(0.2, 0.2);	
	
\draw [fill=gray] (1.7, 0.0) rectangle +(0.2, 0.2);	
\draw [fill=gray] (1.7, 0.2) rectangle +(0.2, 0.2);	
\draw [fill=gray] (1.7, 2.0) rectangle +(0.2, 0.2);			

  \draw (-1.5, 3.2) node {\scriptsize $(M\cdot Z)$};
  \draw (0.3, 3.2) node {\scriptsize $(M\cdot Z)\times N$};
  \draw (1.8, 2.6) node {\scriptsize $N$};
  \draw (-1.5, -.4) node {$\vector x$};
  \draw (-1.1, 1.4) node {$=$};
  \draw (0.3, -.4) node {$\m D$};
  \draw (1.9, -.4) node {$\vector \alpha$};
  \draw (0.3, 0.4) node {\small $\boldmath\ldots$};
  \draw (0.3, 2.4) node {\small $\boldmath\ldots$};
  \draw (-1.5, 1.6) node {\small $\boldmath\vdots$};
  \draw (-0.4, 1.6) node {\small $\boldmath\vdots$};
  \draw (1.0, 1.6) node {\small $\boldmath\vdots$};
  \draw (1.8, 1.3) node {\small $\boldmath\vdots$};
  
\end{tikzpicture}}
\quad
\subfigure[Shapelet-based sparse representation]{
\label{fig:SRc}
\begin{tikzpicture}     
\draw [fill=gray!10] (-1.6, 0) rectangle +(0.2, 2.8);
\foreach \x in {0,...,4}
	\draw [fill=gray] (-1.6, \x/5) rectangle +(0.2, 0.2);
\foreach \x in {0,...,4}
	\draw [fill=gray] (-1.6, \x/5+2.0) rectangle +(0.2, 0.2);
	
\draw [fill=gray!10] (-0.8, 0) rectangle +(2.2, 3.0);
\foreach \x in {0,...,4}
	\draw [fill=be_blue] (-0.8, \x/5) rectangle +(0.2, 0.2);	
\foreach \x in {0,...,4}
	\draw [fill=be_blue] (-0.8, \x/5+2.0) rectangle +(0.2, 0.2);	
\foreach \x in {0,...,4}
	\draw [fill=orange] (-0.6, \x/5) rectangle +(0.2, 0.2);	
\foreach \x in {0,...,4}
	\draw [fill=be_blue] (-0.6, \x/5+2.0) rectangle +(0.2, 0.2);	
\foreach \x in {0,...,4}
	\draw [fill=orange] (-0.4, \x/5) rectangle +(0.2, 0.2);	
\foreach \x in {0,...,4}
	\draw [fill=be_blue] (-0.4, \x/5+2.0) rectangle +(0.2, 0.2);	
\foreach \x in {0,...,4}
	\draw [fill=orange] (-0.2, \x/5) rectangle +(0.2, 0.2);	
\foreach \x in {0,...,4}
	\draw [fill=orange] (-0.2, \x/5+2.0) rectangle +(0.2, 0.2);	
\foreach \x in {0,...,4}
	\draw [fill=orange] (0.6, \x/5) rectangle +(0.2, 0.2);	
\foreach \x in {0,...,4}
	\draw [fill=orange] (0.6, \x/5+2.0) rectangle +(0.2, 0.2);
\foreach \x in {0,...,4}
	\draw [fill=be_blue] (0.8, \x/5) rectangle +(0.2, 0.2);	
\foreach \x in {0,...,4}
	\draw [fill=orange] (0.8, \x/5+2.0) rectangle +(0.2, 0.2);
\foreach \x in {0,...,4}
	\draw [fill=be_blue] (1, \x/5) rectangle +(0.2, 0.2);	
\foreach \x in {0,...,4}
	\draw [fill=orange] (1, \x/5+2.0) rectangle +(0.2, 0.2);
\foreach \x in {0,...,4}
	\draw [fill=orange] (1.2, \x/5) rectangle +(0.2, 0.2);	
\foreach \x in {0,...,4}
	\draw [fill=be_blue] (1.2, \x/5+2.0) rectangle +(0.2, 0.2);
							
\draw [fill=gray!10] (1.7, 0) rectangle +(0.2, 2.4);
\foreach \x in {0,...,3}
	\draw [fill=white] (1.7, \x/5) rectangle +(0.2, 0.2);
\foreach \x in {8,...,11}
	\draw [fill=white] (1.7, \x/5) rectangle +(0.2, 0.2);	

\draw [fill=gray] (1.7, 0.0) rectangle +(0.2, 0.2);	
\draw [fill=gray] (1.7, 0.2) rectangle +(0.2, 0.2);	
\draw [fill=gray] (1.7, 2.0) rectangle +(0.2, 0.2);			

  \draw (-1.5, 3.2) node {\scriptsize $(M\cdot Z)$};
  \draw (0.3, 3.2) node {\scriptsize $(M\cdot Z)\times N$};
  \draw (1.8, 2.6) node {\scriptsize $N$};
  \draw (-1.5, -.4) node {$\vector x$};
  \draw (-1.1, 1.4) node {$=$};
  \draw (0.3, -.4) node {$\m D$};
  \draw (1.9, -.4) node {$\vector \alpha$};
  \draw (0.3, 0.4) node {\small $\boldmath\ldots$};
  \draw (0.3, 2.4) node {\small $\boldmath\ldots$};
  \draw (-1.5, 1.6) node {\small $\boldmath\vdots$};
  \draw (-0.4, 1.6) node {\small $\boldmath\vdots$};
  \draw (1.0, 1.6) node {\small $\boldmath\vdots$};
  \draw (1.8, 1.3) node {\small $\boldmath\vdots$};
  
\end{tikzpicture}}
\label{fig:SR}
\caption{Schematic illustration of sparse representation of an image patch. Colors in the dictionary are indicating the class membership. The pixel-based approach reconstructs all pixels in a patch independently, while the patch-based approach and the shapelet-based approach take spatial information into account by using the image patch in a vectorized way.}
\end{figure*}

In terms of basic sparse coding a $\(V \times 1\)$-dimensional test sample $\vector x$ can be represented by a weighted linear combination of a few elements taken from a $\(V \times N\)$-dimensional dictionary $\m D$, so that $\vector x = \m D \vector \alpha + \vector \epsilon$ with $ \Vert \vector \epsilon \Vert_b$ being the reconstruction error and $b$ specifing the norm for distance computation.
The parameter vector comprising the weights is given by $\vector \alpha$.
The test sample $\vector x$ can be, for example, an $\(M \times 1\)$-dimensional pixel from the image, so that $V=M$, or an $\(\(M\cdot Z\) \times 1\)$-dimensional vectorized image patch $\vector x = \operatorname{vec}(\m X)$ with $Z$ being the number of pixels in the patch and $V=M\cdot Z$.
The reconstruction of an image patch is illustrated in Fig. \ref{fig:SRa} and \ref{fig:SRb}.
While the pixel-based approach reconstructs all pixels in a patch independently,  the patch-based approach takes spatial information into account by using the image patch in a vectorized way.

The optimization problem for determination of optimal $\hat {\vector \alpha}$ is given by
\begin{equation}
	\hat {\vector \alpha} = \operatorname{argmin} \Vert \m D \vector \alpha - \vector x \Vert_b \quad \text{subject to} \quad \Vert \vector \alpha \Vert_{0} \leq W,
\end{equation}
where $W$ is the number of nonzero elements.
Assuming given label information for $N$ dictionary elements $\(\vector d_n, y_n\)$, $n = 1, \ldots, N$, with class labels $y_n \in \mathcal K = \{1, \dots, k, \ldots, K\}$,  the class-wise reconstruction error for $\vector x$ is given by
\begin{equation}
	r_k = \Vert \m D_k \vector \alpha_k -\vector x\Vert_b,
\end{equation}
where $\m D_k$ is a sub-dictionary containing all elements belonging to class $k$.
The test sample $\vector x$ is assigned to the class yielding the lowest reconstruction error.

As mentioned in the introduction, dictionaries are generally designed to be overcomplete, \ie, the number of dictionary elements is larger than the dimension of the dictionary elements.
This is challenging for hyperspectral data due to its high dimensionality, particularly, if spatial information is integrated by using vectorized image patches rather than single pixels.
Moreover, generally, patch-wise label information is limited or not available.
In order to mitigate these challenges, Sec. \ref{sec:dict} shows the construction of an undercomplete, yet powerful dictionary from image patches (see Fig. \ref{fig:SRc}).

As shown in Section \ref{sec:scheme}, under the assumption that an image patch can be factorized into shape and spectral information, the required number of dictionary elements ensuring a successful sparse representation can be reduced significantly.

\subsection{Shapelet-based Image Factorization}
\label{sec:scheme}
\begin{figure}[ht]
	\centering
  	\framebox{\includegraphics[width=0.95\columnwidth]{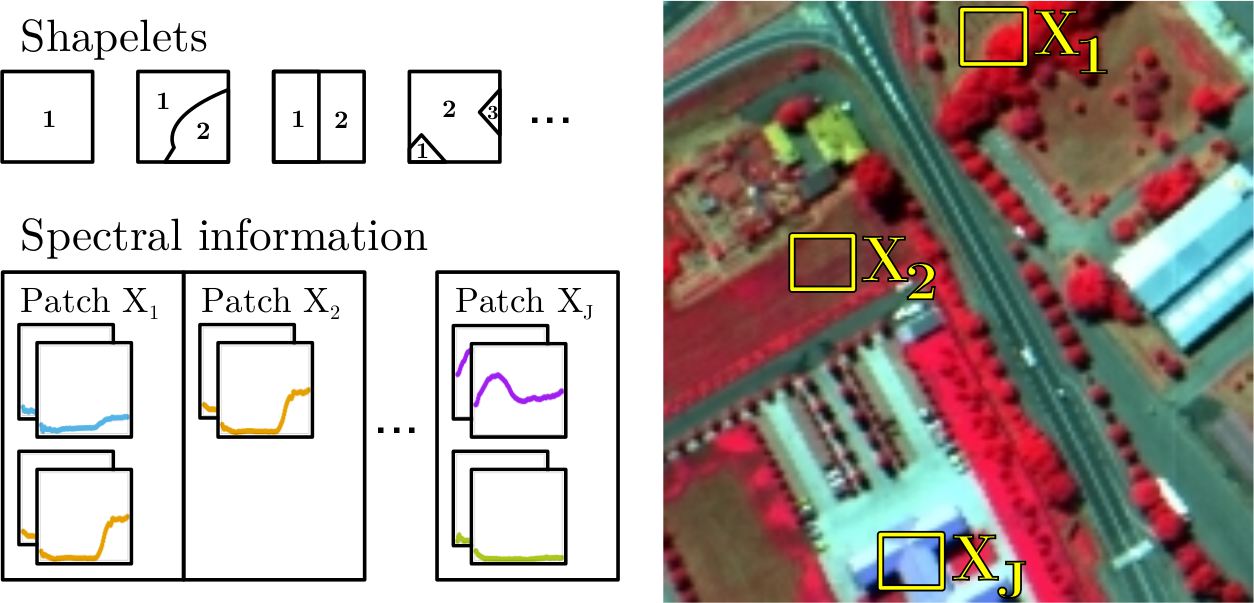}}
  	\caption{Shapelet-based image factorization illustrated by means of a hyperspectral image. Image patches can be described by the composition of a set of learned shapelets with varying number of regions (denoted with numbers) and spectral information, \ie, hyperspectral signatures. In our approach the spectral information is embodied by the training data (different colors of the signatures indicate different classes).}
\label{fig:scheme}
\end{figure}

Fig. \ref{fig:scheme} illustrates the basic idea of the factorization of image patches into shapelets and spectral information, \ie, class-specific hyperspectral signatures.
Dictionaries constructed from patches often conflates shape and spectral information and thus, an unnecessary high amount of dictionary elements are needed for a successful sparse representation \cite{Chua2012}.
Instead of comprising all relevant combinations of shape and spectral information in one global dictionary, the presented approach constructs a specific dictionary for each image patch.
Thus, the dictionary is adequately representative while the number of needed dictionary elements can be significantly reduced, \ie, generally it is not overcomplete.
The basic strategy is to fill each shapelet region with suitable spectral information yielding a highly adaptive dictionary element.
The set of shapelets is learned from each image using a superpixel segmentation approach (see Sec. \ref{sec:shape}) and spectral information are embodied by the training data (see Sec. \ref{sec:spectral}).
For the construction of a patch-specific spatial-spectral dictionary an optimization is performed to choose the best fitting factorization of spectral information and shapelets given a specific image patch (see Sec. \ref{sec:spat_spec}).
Each pixel in the patch is finally classified by a voting scheme explained in Sec. \ref{sec:class}.

\subsection{Dictionary Construction}
\label{sec:dict}

\subsubsection{Shapelet Extraction}
\label{sec:shape}
This step describes the extraction of shapelets from an image.
The $\sqrt{Z} \times \sqrt{Z}$-dimensional shapelets $\m S_n$ with $n \in \{1, \dots, N\}$, each containing $R_n$ regions, are specifically learned for each image using a superpixel segmentation approach.
They constitute the most representative repeated patterns, \ie, characteristic local arrangement of  pixels with similar features.
Here one region indicates an area of homogeneous features.

\begin{figure}
\centering
	\includegraphics[width=0.5\columnwidth]{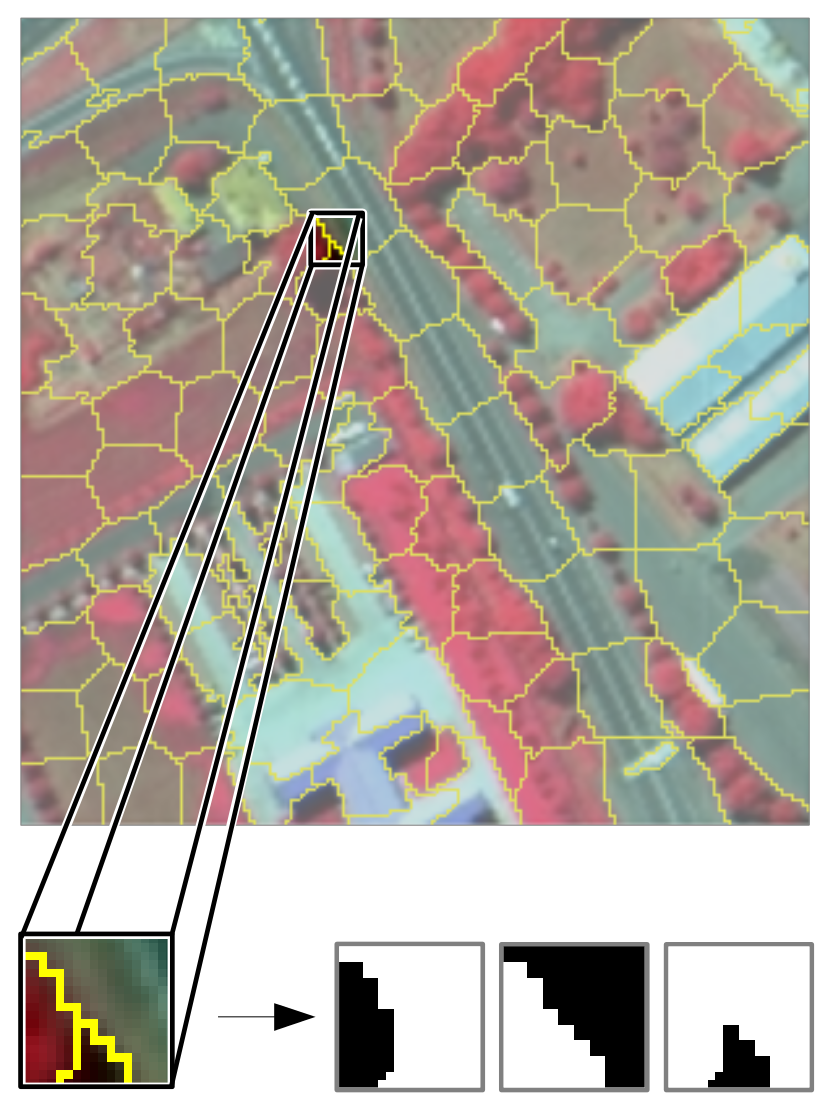}
	\label{fig:shapelets}
	\caption{Superpixel segmentation and by way of illustration one conversion from an image patch to binary patches in order to find representative shapelets by clustering. The binary cluster centers are indexed and constitute the shapelets.}
\end{figure}

\begin{figure}
\centering
	\includegraphics[width=0.4\columnwidth]{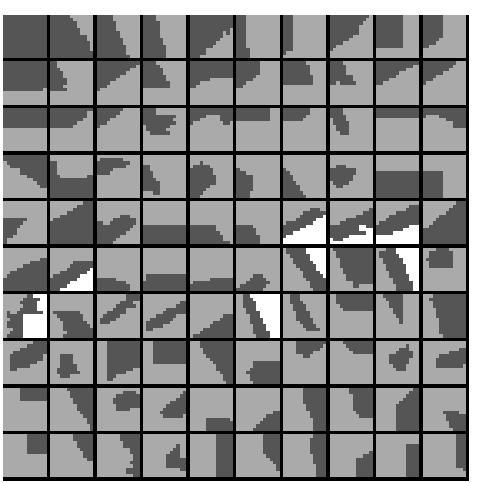}
	\includegraphics[width=0.32\columnwidth]{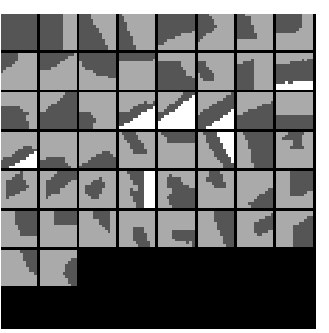}
	\includegraphics[width=0.2\columnwidth]{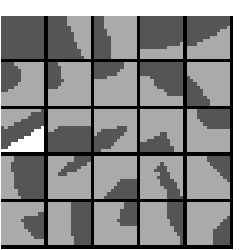}
	\caption{Visualized shapelet sets with 100, 50 and 25 shapelets, where different colors indicate the shapelet regions. }
	\label{fig:shapelets_results}
\end{figure}

Fig. \ref{fig:shapelets} illustrates the extraction of image-specific shapelets.
First, the simple linear iterative clustering (SLIC) superpixel segmentation approach \cite{Achanta2012} is applied to the z-normalized image using all image bands.
It provides a segmentation of the image into compact segments each of them containing homogeneous spectral information.
Given a binary segmentation mask for the image (illustrated as yellow lines in Fig. \ref{fig:shapelets}), overlapping image patches of size $\sqrt{Z} \times \sqrt{Z}$ are extracted and converted to binary patches.
In detail, for each region a binary patch is created with the region's pixels set to 0 and the remaining pixel to 1.
The most representative segmentation patterns are found by a k-medoids clustering \cite{Kaufman1987}, where the cluster centers are reshaped to $\sqrt{Z} \times \sqrt{Z}$ and constitute the shapelets (see Fig. \ref{fig:shapelets_results}). 
The characteristic of the dictionary is influenced by the number of cluster centers and the approximate size and compactness of the segments yielded by the SLIC superpixel algorithm. 

\subsubsection{Spectral Information Choice}
\label{sec:spectral}
In the presented approach the spectral information is assumed to be the training data.
The training set consists of $L$ labeled samples $\({}^\text{L}{\vector x_l}, {}^\text{L}y_l\)$ with $ l \in \{1, \ldots, L\}$ comprising $M$-dimensional spectral feature vectors ${}^\text{L}{\vector x_l} \in \mR^M$ and class labels ${}^\text{L}{\vector y} = \left[{}^\text{L}y_l\right]$ with ${}^\text{L}y_l \in \{1, \dots, k, \ldots, K\}$. 
The feature vectors are collected in an ($M\times L$)-dimensional matrix ${}^\text{L}\m X  = \left[{}^\text{L}{\vector x_1}, \ldots, {}^\text{L}{\vector x_L}\right]$, where the upper left index $\text{L}$ refers to the term \emph{labeled data}. 

\subsubsection{Patch-Specific Spatial-Spectral Dictionary Construction}
\label{sec:spat_spec}
\begin{figure*}
\centering
	\subfigure{\begin{tikzpicture}[scale=0.9]
\begin{scope}[
           xshift=0.1,yshift=3, every node/.append style={
           yslant=0.5,xslant=-1},yslant=0.5,xslant=-1
           ]
       \draw[step=4mm, black] (0,0) grid (3.2,3.2); 
       \draw[black,thick] (0,0) rectangle (3.2,3.2);
       \draw[red,thick] (1.2,1.6) rectangle (2.4,2.8);
       \fill[brown, fill opacity=1] (2.05,2.05) rectangle (2.35,2.35); 
       \fill[brown] (1.65,2.05) rectangle (1.95,2.35);
       \fill[brown] (2.05,2.45) rectangle (2.35,2.75);
       \fill[brown] (1.65,2.45) rectangle (1.95,2.75);
       \fill[brown] (1.25,2.45) rectangle (1.55,2.75);
       \fill[blue] (2.05,1.95) rectangle (2.35,1.65);
       \fill[blue] (1.65,1.95) rectangle (1.95,1.65); 
       \fill[blue] (1.25,2.05) rectangle (1.55,2.35);
       \fill[blue] (1.25,1.65) rectangle (1.55,1.95);
   \end{scope}
   \node[text=black] at (3.5,3) {Shapelet-based layer};
   \node[text=black] at (3.5,-0.8) {Pixel-based layer};
   \begin{scope}[
           xshift=0,yshift=-100, every node/.append style={
           yslant=0.5,xslant=-1},yslant=0.5,xslant=-1
           ]
       \fill[white,fill opacity=0.1] (0,0) rectangle (3.2,3.2); 
       \draw[step=4mm, black] (0,0) grid (3.2,3.2); 
       \draw[black,thick] (0,0) rectangle (3.2,3.2);
       \draw[red,thick] (1.2,1.6) rectangle (2.4,2.8);
       \fill[gray] (2.05,2.45) rectangle (2.35,2.75);
       \fill[gray] (1.65,2.45) rectangle (1.95,2.75);
       \fill[gray] (1.25,2.45) rectangle (1.55,2.75);
       \fill[gray] (2.05,2.05) rectangle (2.35,2.35);
       \fill[gray] (1.65,2.05) rectangle (1.95,2.35);
       \fill[gray] (1.25,2.05) rectangle (1.55,2.35);
       \fill[gray] (2.05,1.65) rectangle (2.35,1.95);
       \fill[gray] (1.65,1.65) rectangle (1.95,1.95);
       \fill[gray] (1.25,1.65) rectangle (1.55,1.95);
  \end{scope}
\end{tikzpicture}}
	\qquad
	\subfigure{
	\begin{tikzpicture}[scale=1]
\begin{scope}[rotate around = {-0:(0,0,0)}]
\shadedraw [ball color = brown!150] (0.75,2,1.5) circle (0.25);			
\shadedraw [ball color = blue!95] (2.25,2,1.5) circle (0.25); 	
	    	
\shadedraw [ball color = gray!10] (0,0,0) circle (0.25);
\shadedraw [ball color = gray!10] (0,0,1.5) circle (0.25);
\shadedraw [ball color = gray!10] (0,0,3) circle (0.25);
\shadedraw [ball color = gray!10] (1.5,0,0) circle (0.25); 	
\shadedraw [ball color = gray!10] (1.5,0,1.5) circle (0.25);
\shadedraw [ball color = gray!10] (1.5,0,3) circle (0.25);
\shadedraw [ball color = gray!10] (3,0,0) circle (0.25);
\shadedraw [ball color = gray!10] (3,0,1.5) circle (0.25);  	
\shadedraw [ball color = gray!10] (3,0,3) circle (0.25);
    	
\shadedraw [ball color = black!80] (0,-2,0) circle (0.25);
\shadedraw [ball color = black!80] (0,-2,1.5) circle (0.25);
\shadedraw [ball color = black!80] (0,-2,3) circle (0.25);
\shadedraw [ball color = black!80] (1.5,-2,0) circle (0.25);
\shadedraw [ball color = black!80] (1.5,-2,1.5) circle (0.25);
\shadedraw [ball color = black!80] (1.5,-2,3) circle (0.25);
\shadedraw [ball color = black!80] (3,-2,0) circle (0.25);
\shadedraw [ball color = black!80] (3,-2,1.5) circle (0.25);    	
\shadedraw [ball color = black!80] (3,-2,3) circle (0.25);
\shadedraw [ball color = black!80] (2.25,3,1.5) circle (0.25);
\shadedraw [ball color = black!80] (0.75,3,1.5) circle (0.25); 

\shadedraw [ball color = black!80, opacity=0.1] (0,-2,0) circle (0.25);
\shadedraw [ball color = black!80, opacity=0.1] (0,-2,1.5) circle (0.25);
\shadedraw [ball color = black!80, opacity=0.1] (0,-2,3) circle (0.25);
\shadedraw [ball color = black!80, opacity=0.1] (1.5,-2,0) circle (0.25);
\shadedraw [ball color = black!80, opacity=0.1] (1.5,-2,1.5) circle (0.25);
\shadedraw [ball color = black!80, opacity=0.1] (1.5,-2,3) circle (0.25);
\shadedraw [ball color = black!80, opacity=0.1] (3,-2,0) circle (0.25);
\shadedraw [ball color = black!80, opacity=0.1] (3,-2,1.5) circle (0.25);    	
\shadedraw [ball color = black!80, opacity=0.1] (3,-2,3) circle (0.25);
\shadedraw [ball color = black!80, opacity=0.1] (2.25,3,1.5) circle (0.25);
\shadedraw [ball color = black!80, opacity=0.1] (0.75,3,1.5) circle (0.25);

\path [very thick] (0.75,2.25,1.5) edge node {} (0.75,2.75,1.5);
\path [very thick] (2.25,2.25,1.5) edge node {} (2.25,2.75,1.5);

\path [very thick] (0,-0.25,0) edge node {} (0,-1.75,0);
\path [very thick] (0,-0.25,1.5) edge node {} (0,-1.75,1.5);
\path [very thick] (0,-0.25,3) edge node {} (0,-1.75,3);
\path [very thick] (1.5,-0.25,0) edge node {} (1.5,-1.75,0);
\path [very thick] (1.5,-0.25,1.5) edge node {} (1.5,-1.75,1.5);
\path [very thick] (1.5,-0.25,3) edge node {} (1.5,-1.75,3);
\path [very thick] (3,-0.25,0) edge node {} (3,-1.75,0);
\path [very thick] (3,-0.25,1.5) edge node {} (3,-1.75,1.5);
\path [very thick] (3,-0.25,3) edge node {} (3,-1.75,3);

\path [very thick] (0.75,1.75,1.5) edge node {} (0,0.25,0);
\path [very thick] (0.75,1.75,1.5) edge node {} (1.5,0.25,0);
\path [very thick] (0.75,1.75,1.5) edge node {} (0,0.25,1.5);
\path [very thick] (0.75,1.75,1.5) edge node {} (0,0.25,3);
\path [very thick] (0.75,1.75,1.5) edge node {} (1.5,0.25,1.5);
\path [very thick] (2.25,1.75,1.5) edge node {} (3,0.25,0);
\path [very thick] (2.25,1.75,1.5) edge node {} (3,0.25,1.5);
\path [very thick] (2.25,1.75,1.5) edge node {} (3,0.25,3);
\path [very thick] (2.25,1.75,1.5) edge node {} (1.5,0.25,3);
	
\end{scope}
\end{tikzpicture}}
	\caption{Graphical model used for determination of the best fitting spectral information given the shapelet regions. Exemplarily, one shapelet is illustrated (red border) overlaying the image layer. Left: The lower, pixel-based layer contains the pixel information (illustrated in gray) and the upper layer contains the shapelet regions (illustrated in brown and blue). Right: Dark nodes denote feature nodes, light gray nodes denote pixel-wise class label nodes and the brown and orange nodes denote the class label nodes for the shapelet regions. Each label node is connected to a feature node and class label nodes for the shapelet regions are connected to all class label nodes for all pixels lying in the shapelet region.}
	\label{fig:graph}
\end{figure*}
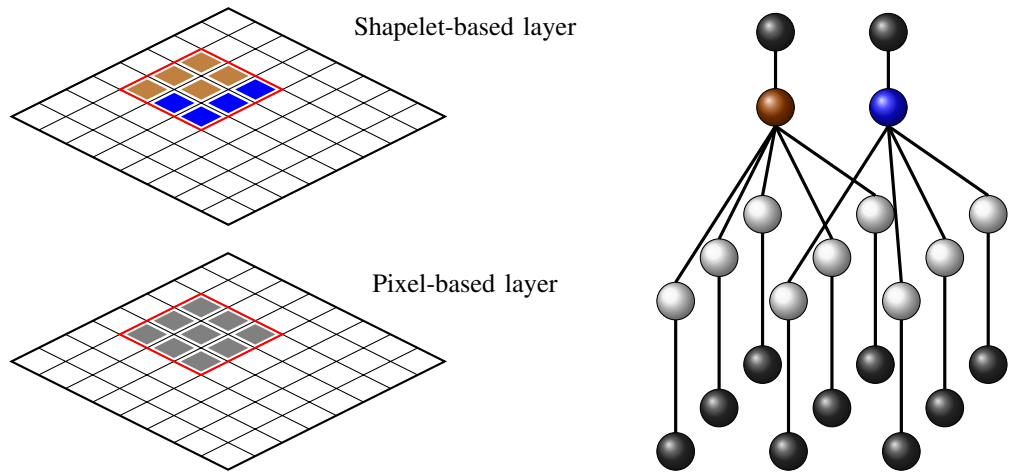

In this step shapelets and spectral information are combined to a $\(\(Z \cdot M\) \times N\)$-dimensional patch-specific spatial-spectral dictionary $\m D_j$ for each patch $\m X_j$.
A schematic overview and the algorithm is provided in Fig. \ref{fig:algorithm}.
As illustrated in Fig. \ref{fig:graph}, a hierarchical Markov random field (MRF) is employed for the construction of the $n$-th element in the spatial-spectral dictionary $\vector d_{j,n}$. 
In this way, prior knowledge about the relation between pixels and shapelet regions is modeled, where the lower, pixel-wise layer contains pixel information and the upper layer contains shapelet region information.
In order to keep the notation uncluttered, the indices $(\cdot)_j$ and $(\cdot)_n$ are omitted for further considerations in this paragraph, although the following procedure is repeated for all shapelets and all image patches.
The MRF determines the indices ${}^\text{P}\vector b = \left[{}^\text{P} b_z\right]$, $z=1,\ldots, Z$, with ${}^\text{P} b_z \in \{1, \dots, L\}$, and ${}^\text{S}\vector b = \left[{}^\text{S} b_r\right]$, $r=1,\ldots, R$, with ${}^\text{S} b_r \in \{1, \dots, L\}$, of the best fitting spectral information for both pixel-based layer (indicated by upper left index~$\text{P}$) and shapelet-based layer (indicated by upper left index~$\text{S}$). 
More specifically, given an image patch the MRF is used to determine the indices of the training samples which are used to fill the shapelet.

For this, an energy function is minimized utilizing message passing \cite{MacKay2003}:
\begin{align}
	\label{eq:energy}
	E({}^\text{P}\vector b, {}^\text{S}\vector b) &= -\sum_z \operatorname{corr} \left(\vector x_z, \ical x \left({}^\text{P} b_z\right) \right) - 
	\gamma \sum_r \ical h\left(\ical y \left({}^\text{S} b_r\right)\right) + \nonumber\\
	& \hspace{1.5em}
	\omega \sum_r \sum_{z \in \mathcal S_r} \delta \left(\ical y \left({}^\text{P} b_z\right), \ical y \left({}^\text{S} b_r\right) \right),
\end{align}
where $\vector x_z$ is the $z$-th pixel in the image patch and $\ical x \left(\cdot\right)$, $\ical y \left(\cdot\right)$, $\ical h \left(\cdot\right)$ are operators which access elements at position $\left(\cdot\right)$ in ${}^\text{L}\m X$, ${}^\text{L}\vector y$ and $\vector h_r$ (see Eq. \ref{eq:hist}), respectively. 
Indices within the $r$-th region in shapelet $\m S$ are denoted with $z \in \mathcal S_r$, $\delta\(\cdot, \cdot\)$ is the delta function, and $\omega$ and $\gamma$ are weighting factors between both unary terms and the third, binary term.
The second unary term uses the normalized histogram $\vector h_r$ over a rough estimation of class labels $\widetilde{\vector y}_r = \left[\widetilde{y}_z\right]$, $z \in \mathcal S_r$, calculated as 
\begin{align}
	\label{eq:hist}
	\vector h_{r} & = \left[ h_1, \ldots, h_k, \ldots, h_K \right] = \frac{1}{\left| \mathcal S_r \right|} \operatorname{hist}\(\widetilde{\vector y}_r\), \\
	\label{eq:hist2}
	\widetilde{y}_z & = \ical y \(\operatorname{argmax}_l\(\operatorname{corr} \left(\vector x_z,{}^\text{L} \vector x_l \right)\)\),
\end{align}
with $\left| \mathcal S_r \right|$ being the number of elements in $r$-th region of a shapelet.
Here, the estimated class labels are determined by a simple nearest neighbor classifier.
The usage of this term supports the selection of spectral information belonging to the dominant class, \ie, the class that would be chosen by majority vote.
However, also other kind of unary and binary terms can be chosen. 
The final dictionary element $\vector d$ consists of elements $\ical x \left({}^\text{P} \vector b\right)$ and the dictionary element labels are given by $\ical y \left({}^\text{P} \vector b\right)$.

\begin{center}
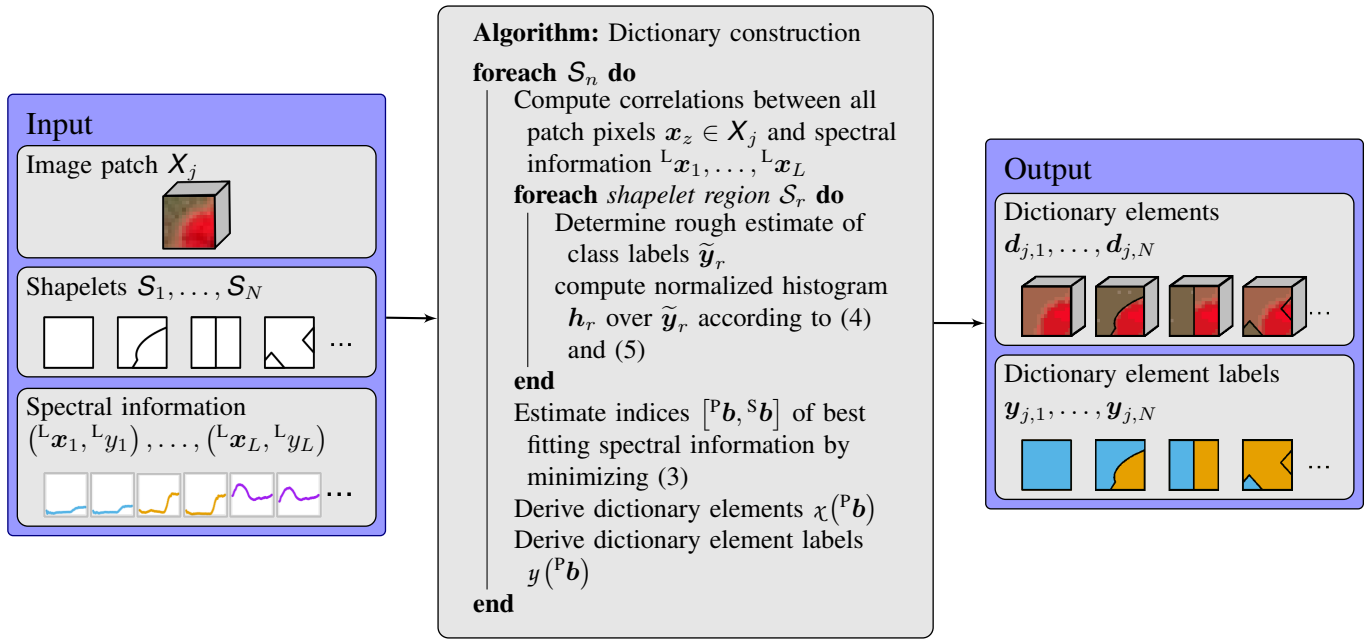
\begin{figure*}
\tikzstyle{block} = [rectangle, draw, fill=gray!20, 
    text width=0.25\textwidth, rounded corners] 
\tikzstyle{line} = [draw, -latex']
     
\begin{tikzpicture}[[
    every path/.style={line width=0.5pt,rounded corners=0.2em},
    every path/.style={line width=0.5pt,rounded corners=0.2em},
    circ/.style={draw=blue!50!black,thick,circle,fill=white,minimum width=0.25\textwidth},
    link/.style={thick,->},node distance = 2cm, auto]
    
    \node [rectangle, draw=none, text width=0.25\textwidth] (input) {\large Input};
    \node [block, below of=input, node distance=1cm] (init) {Image patch $\m X_j$
    \begin{center} 
    \includegraphics[width=0.2\textwidth]{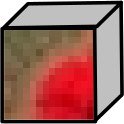}
    \end{center}};
    
    \node [block, below of=init, node distance=1.6cm] (shapelets) {Shapelets $\m S_1, \ldots, \m S_N$	
    \begin{center}
    \includegraphics[width=0.9\textwidth]{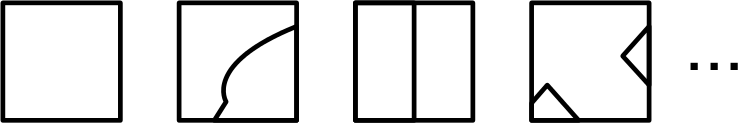}
    \end{center}};
    
    \node [block, below of=shapelets, node distance=1.8cm] (spectral) {Spectral information $\left({}^\text{L}{\vector x_1}, {}^\text{L}y_1\right), \ldots, \left({}^\text{L}{\vector x_L}, {}^\text{L}y_L\right)$
    \begin{center}
    \includegraphics[width=0.9\textwidth]{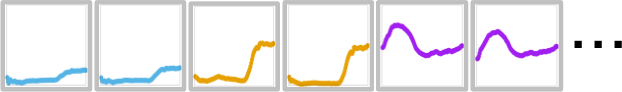}
    \end{center}};

	\node [block, right of=shapelets, node distance=6.5cm,fill=gray!20, text width=0.35\textwidth] (construction) {
 		\removelatexerror
 		\begin{algorithm}[H]
		\DontPrintSemicolon
		\TitleOfAlgo{Dictionary construction}
 		\ForEach{$\m S_n$}{
  Compute correlations between all patch pixels $\vector x_z\in \m X_j$ and spectral information ${}^\text{L}{\vector x_1}, \ldots, {}^\text{L}{\vector x_L}$\;
  		\ForEach{shapelet region $\mathcal S_r$}{
  Determine rough estimate of class labels $\widetilde{\vector y}_r$\;
  compute normalized histogram $\vector h_{r}$ over $\widetilde{\vector y}_r$ according to (\ref{eq:hist}) and (\ref{eq:hist2})\;}
  Estimate indices $\left[{}^\text{P}\vector b, {}^\text{S}\vector b\right]$ of best fitting spectral information by minimizing (\ref{eq:energy})\;
  Derive dictionary elements $\ical x \left({}^\text{P} \vector b\right)$\;
  Derive dictionary element labels $\ical y \left({}^\text{P} \vector b\right)$\;
 }
		\end{algorithm}
};
 
    \node [rectangle, right of=construction, draw=none, text width=0.25\textwidth, minimum height=1em, node distance=6.5cm, yshift=2cm] (output) {\large Output};
    \node [block, below of=output, node distance=1.3cm] (result) {Dictionary elements $\vector d_{j,1}, \ldots, \vector d_{j,N}$
    \begin{center} 
    \includegraphics[width=0.9\textwidth]{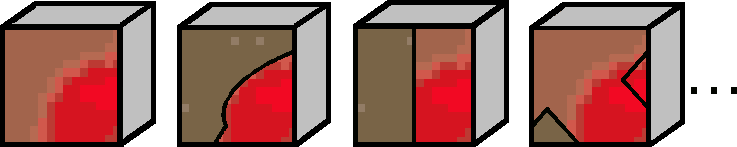}
    \end{center}};
    
    \node [block, below of=result, node distance=2.1cm] (result2) {Dictionary element labels $\vector y_{j,1}, \ldots, \vector y_{j,N}$
    \begin{center} 
    \includegraphics[width=0.9\textwidth]{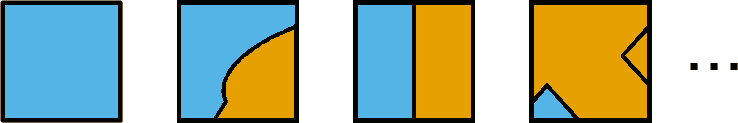}
    \end{center}};

  \begin{pgfonlayer}{background}
    \node [draw,fit=(input)(init)(shapelets)(spectral),label={left:},draw=blue!50!black,fill=blue!40] (group1) {};
    \node [draw,fit=(output)(result)(result2),label={left:},fill=blue!40] (group2){};
  \end{pgfonlayer}

   \path [line, thick] (group1) -- (construction);
   \path [line, thick] (construction) -- (group2);
  	
\end{tikzpicture}
\caption{Schematic overview of the construction of the patch-specific spatial-spectral dictionary.}
\label{fig:algorithm}
\end{figure*}
\end{center}

\subsection{Classification}
\label{sec:class}
Given the spatial-spectral dictionary $\m D_j$, the classification of an image can be derived by performing the sparse coding procedure for each image patch $\m X_j$ with
\begin{equation}
	\label{eq:sparse}
	\hat {\vector \alpha_j} = \operatorname{argmin} \Vert \m D_j \vector \alpha_j - \vector x_j \Vert_b \qquad \text{subject to} \qquad \Vert \vector \alpha_j \Vert_{0} \leq W.
\end{equation}
The optimization is done via orthogonal matching pursuit.

Since the image patches are fully overlapping, a voting scheme is introduced in order to derive the final label of each pixel by a weighted voting. 
The votes for class $k$ of the $z$-th pixel in $j$-th patch $x_{j,z}$ can be obtained by $v^k_{j,z} = 1 / r^k_{j,z}$, where $r^k_{j,z}$ is the reconstruction error given by
\begin{equation}
	r^k_{j,z}= \Vert x_{j,z} - \m D^k_{j,z} \hat{\vector \alpha}^k_{j,z}\Vert
	\label{eq:voting}
\end{equation}
with $\m D^k_{j,z}$ being a sub-matrix of the whole dictionary $\m D_j$.
The rows of the sub-matrix are chosen according as they are necessary to reconstruct the $z$-th pixel and the columns are chosen according to the class membership. 
The parameters $\hat{\vector \alpha}^k_{j,z}$ are the ones who are assigned to the dictionary elements $\m D^k_{j,z}$.
All votes assigned to a test pixel are summed up and the class with the highest vote is chosen.

\section{Experimental Setup and Datasets}
\label{sec:experiments}
In this section the experimental setup is presented in order to evaluate our proposed sparse representation-based classification approach with spatial-spectral dictionary by means of the overall accuracy, average accuracy and $\kappa$ coefficient.

\subsection{Datasets}
The considered datasets are three widely used hyperspectral images - \textsc{Indian Pines}, \textsc{University of Pavia} and \textsc{Center of Pavia} - from study sites with different environmental setting.
The \textsc{Indian Pines} dataset was acquired by the AVIRIS instrument in 1992. 
The study site lies in a predominately agricultural region in NW Indiana, USA.
The dataset covers $145 \times 145$ pixels, with a spatial resolution of $20$~m per pixel.
Some bands covering water absorption, \ie~$\left[104-108\right]$, $\left[150-163\right]$ and $220$, were removed resulting in $200$ spectral bands.
The training data is randomly selected and comprises about $10\%$ of the labeled data (see Fig.~\ref{fig:results} and Tab.~\ref{tab:results_indian}). 
The \textsc{University of Pavia} dataset was also acquired by ROSIS-3 sensor with $610 \times 340$ pixels in size and $103$ channels. 
The classification is aiming on nine land cover classes. 
Fig.~\ref{fig:results} and Tab.~\ref{tab:results_paviaUni} show the training data.
The \textsc{Center of Pavia} image was acquired by ROSIS-3 sensor in 2003 with a spatial resolution of $1.3$~m per pixel. 
Some bands have been removed due to noise, and finally $102$ channels have been used in the classification. 
The image strip, with $1096 \times 492$ pixels in size, lies around the center of Pavia. 
The classification is aiming on nine land cover classes. 
Fig. \ref{fig:results} and Tab. \ref{tab:results_paviaCenter} show the training data, which covers about 9\% of the whole data.
All images are z-normalized.

\subsection{Experimental Setup}
Each image is sparsely represented using the methods presented in Sec. \ref{sec:methods} and classified using (\ref{eq:voting}).
Orthogonal matching pursuit is used to solve the sparse coding task in (\ref{eq:sparse}).
In the experiments the number of shapelets, the number of used dictionary elements, the patchsize and the superpixel size are varied (see Fig. \ref{fig:numshape}, Fig. \ref{fig:numW} and Fig. \ref{fig:SPsize}).
The best parameter setting is chosen in order to compare our proposed approach (\name) to support vector machines with composite kernel (SVMCK, \cite{Camps-Valls2006}), the joint sparsity model using simultaneous orthogonal matching pursuit (SOMP), simultaneous subspace pursuit (SSP), sparse coding approach
with spectral-contextual dictionary learning (SCDL, \cite{Soltani-Farani2013}) and the three best performing kernelized sparse coding algorithms presented in \cite{Chen2013}: kernel subspace pursuit with composite kernel (KSPCK), kernel simultaneous subspace pursuit (KSSP) and kernel orthogonal matching pursuit with composite kernel (KOMPCK).
All results are taken from \cite{Chen2013} except SCDL, which is taken from \cite{Soltani-Farani2013}.
Accuracy assessment was performed with independent test data, giving overall accuracies, average accuracies, $\kappa$ coefficient, and confusion matrices that were used to calculate the class accuracies. 
The training and test data sets as well as the sampling scheme used for the proposed \name~approach are identical to \cite{Chen2013} and \cite{Soltani-Farani2013} to ensure the comparability of the results.

\section{Experimental Results and Discussion}
\label{sec:results}
\subsection{\textsc{Indian Pines} Dataset}
The averaged accuracy measures for ten results with randomly sampled training and test data, achieved by using 10 shapelets and a patchsize of $9 \times 9$ are shown in Tab. \ref{tab:results_indian}.
The training data, test data and the classification map of the run with the highest average accuracy is shown in Fig. \ref{fig:results}.

\begin{table*}[ht]
\tabcolsep2pt
\centering
\footnotesize
\caption{Size of training and test data, classwise accuracies [\%], overall accuracy [\%], average accuracy [\%] and $\kappa$ coefficient of \textsc{Indian Pines} dataset. Numbers in brackets show the standard deviation over 10 randomly sampled training- and testsets. Bold numbers indicate the best results. Bold number indicates the best results and underlined results the second best result.}
\begin{tabular}{lcccccccccc}
	& \# train & \# test & SVMCK & SOMP & SSP & KSPCK & KSSP & KOMPCK & SCDL & \name\\
	\hline
\colorbox{i1}{\textcolor{i1}{O}} \textsc{Alfalfa} & 6 & 48 & \textbf{95.83} & 85.42 &  81.25 & \underline{95.83} & 91.67 & \textbf{97.92} & 93.75 & \underline{95.83} (3.30)\\
\colorbox{i2}{\textcolor{i2}{O}} \textsc{Corn-notill} & 144 & 1290 & 96.67 & 94.88 & 95.74 & \underline{99.15} & 97.98 & \textbf{99.22} &  94.93 & 97.95 (1.67)\\
\colorbox{i3}{\textcolor{i3}{O}} \textsc{Corn-min} & 84 & 750 & 90.93 & 94.93 & 92.80 & 96.93 & \underline{97.73} & 96.93 &  97.39 & \textbf{98.56} (2.72)\\
\colorbox{i4}{\textcolor{i4}{O}} \textsc{Corn}& 24 & 210 	& 85.71 & 91.43  & 82.38 & \underline{97.14} & 96.67 & 95.24 &  90.57 & \textbf{98.38} (1.74)\\
\colorbox{i5}{\textcolor{i5}{O}} \textsc{Grass/Pasture} 	& 50 & 447 	& 93.74 & 89.49 & 93.29 & 98.21 & 94.85 & \underline{98.43} &  97.23 & \textbf{99.10} (0.91)\\
\colorbox{i6}{\textcolor{i6}{O}} \textsc{Grass/Trees} & 75 & 672 & 97.32 	& 98.51 & 98.81 & 99.11 & 98.96 & 99.11 &  \underline{99.17} & \textbf{99.70} (0.28)\\
\colorbox{i7}{\textcolor{i7}{O}} \textsc{Grass/Pasture-mowed} & 3 & 23 & 69.57 & 91.30 & 82.61 & \textbf{100.00} & 17.39 & \textbf{100.00} & \textbf{100.00} & \underline{96.52} (5.67)\\
\colorbox{i8}{\textcolor{i8}{O}} \textsc{Hay-windrowed} & 49 & 440 & 98.41 & 99.55 & 99.77 & \underline{99.97} & \textbf{100.00} & \textbf{100.00} & 99.95 & 99.95 (0.10)\\
\colorbox{i9}{\textcolor{i9}{O}} \textsc{Oats} & 2 & 18 & 55.56 & 0.00  & 0.00 & \textbf{100.00} & 0.00 & 88.89 &  79.44 & \underline{95.55} (6.09) \\
\colorbox{i10}{\textcolor{i10}{O}} \textsc{Soybeans-notill} & 97 & 871		& 93.80 & 89.44 & 91.27 & 97.70 & 94.37 & \underline{98.05} &  96.30 & \textbf{98.62} (0.86)\\
\colorbox{i11}{\textcolor{i11}{O}} \textsc{Soybeans-min} & 247 & 2221 		& 94.37 & 97.34  & 97.43 & 98.20 & 98.33 & 97.43 &  \underline{98.46} & \textbf{99.23} (0.92)\\
\colorbox{i12}{\textcolor{i12}{O}} \textsc{Soybeans-clean} & 62 & 552 		& 93.66 & 88.22 & 89.13 & \textbf{98.73} & 97.46 & \textbf{98.73} &  92.97 & \underline{97.57} (1.44)\\
\colorbox{i13}{\textcolor{i13}{O}} \textsc{Wheat} & 22 & 190 		& 99.47 & \textbf{100.00} & 99.47 & \textbf{100.00} & \textbf{100.00} & \textbf{100.00} &  99.05 & \underline{99.68} (0.71)\\
\colorbox{i14}{\textcolor{i14}{O}} \textsc{Woods} & 130 & 1164 	& 99.14 & 99.14 & 99.05 & \underline{99.48} & \textbf{99.91} & 99.40 &  98.87 & \textbf{99.91} (0.15)\\
\colorbox{i15}{\textcolor{i15}{O}} \textsc{Building-Grass-Trees} & 38 & 342 		& 87.43 & \textbf{99.12} & 97.95 & 97.37 & 97.08 & 97.95 &  97.13 & \underline{98.71} (1.12)\\
\colorbox{i16}{\textcolor{i16}{O}} \textsc{Stone-steel Towers} & 10 & 85 		& \textbf{100.00} & 96.47 & 92.94 & 95.29 & 94.12 & \underline{97.65} &  96.00 &  92.94 (4.63)\\
\hline
Overall 			& 1043 & 9323 	& 94.86  & 95.28 & 95.34 & \underline{98.47} & 97.46 & 98.33 &  97.81 & \textbf{98.87} (0.19)\\
Average 	    		&  & 			& 90.73  & 88.45 & 87.12 & \textbf{98.31} & 86.03 & 97.81 &  95.70 & \underline{98.02} (0.76)\\
$\kappa$ 		& & 				& 0.941  & 0.946 & 0.947 & \underline{0.983} & 0.971 & 0.981 &  0.964 & \textbf{0.987} (0.002)\\
\end{tabular}
\label{tab:results_indian}
\end{table*}
       
The results show that our proposed approach achieves higher classification accuracies than SVMCK, SOMP, SSP and SCDL.
It achieves comparable results to kernel subspace pursuit with composite kernel (KSPCK). 
Regarding the class-specific accuracies our proposed approach performs well, resulting in the highest or second-best accuracy in the very most cases. 
Even small classes (\eg, \textsc{Oats}) are accurately classified.
While most classes are classified stable, resulting in low standard deviations, the accuracies of small classes (\eg, \textsc{Oats} and \textsc{Grass/Pasture-mowed}) show relatively high standard deviations. 
Thus, as for other classifiers, the selection of adequate training samples is important.

\begin{figure*}[ht]
  \centering
	\framebox{\subfigure[Overall accuracy \textsc{Indian Pines}]{\includegraphics[width=0.45\textwidth]{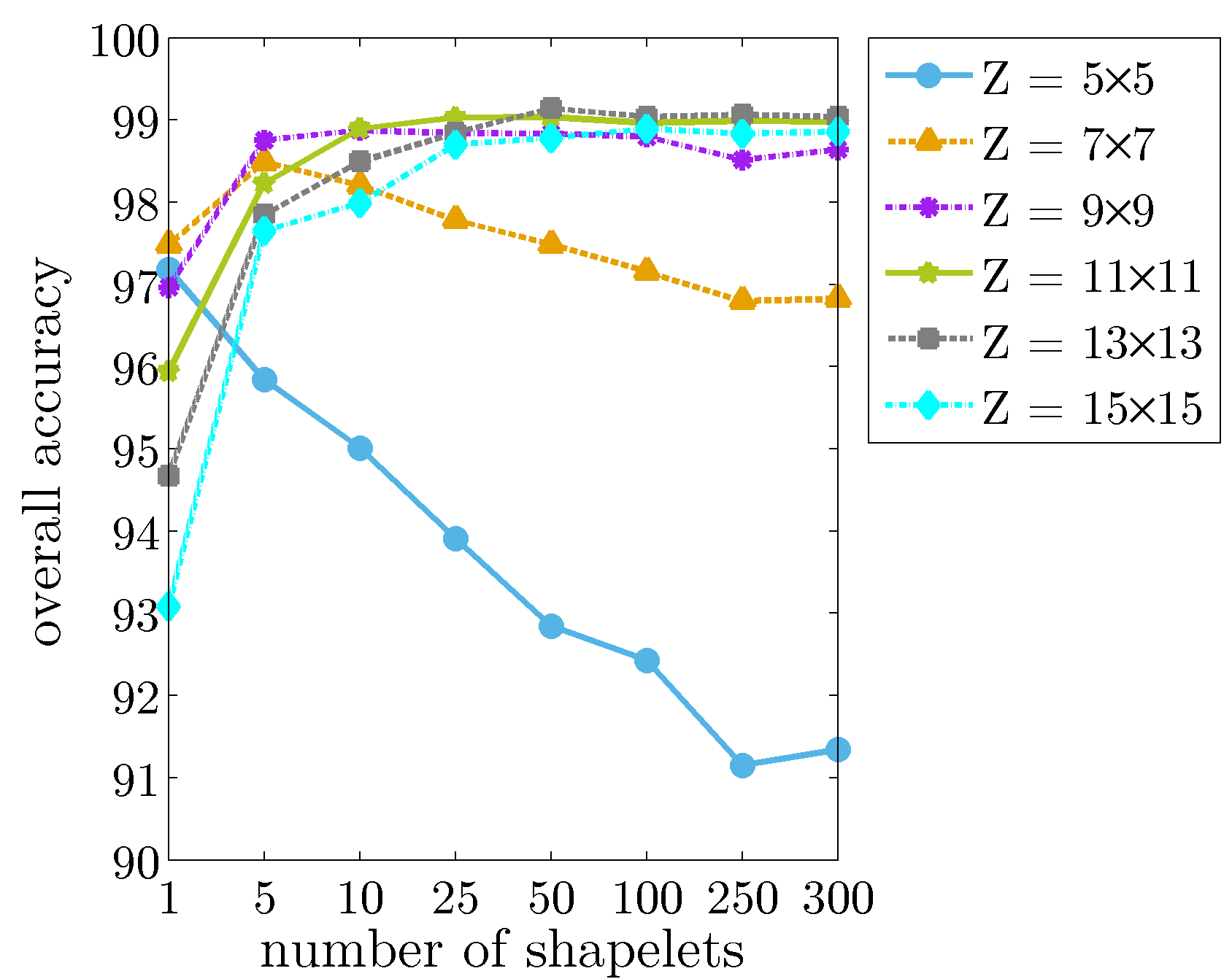}\label{fig:indian_numshape1}}}
	\framebox{\subfigure[Average accuracy \textsc{Indian Pines}]{\includegraphics[width=0.45\textwidth]{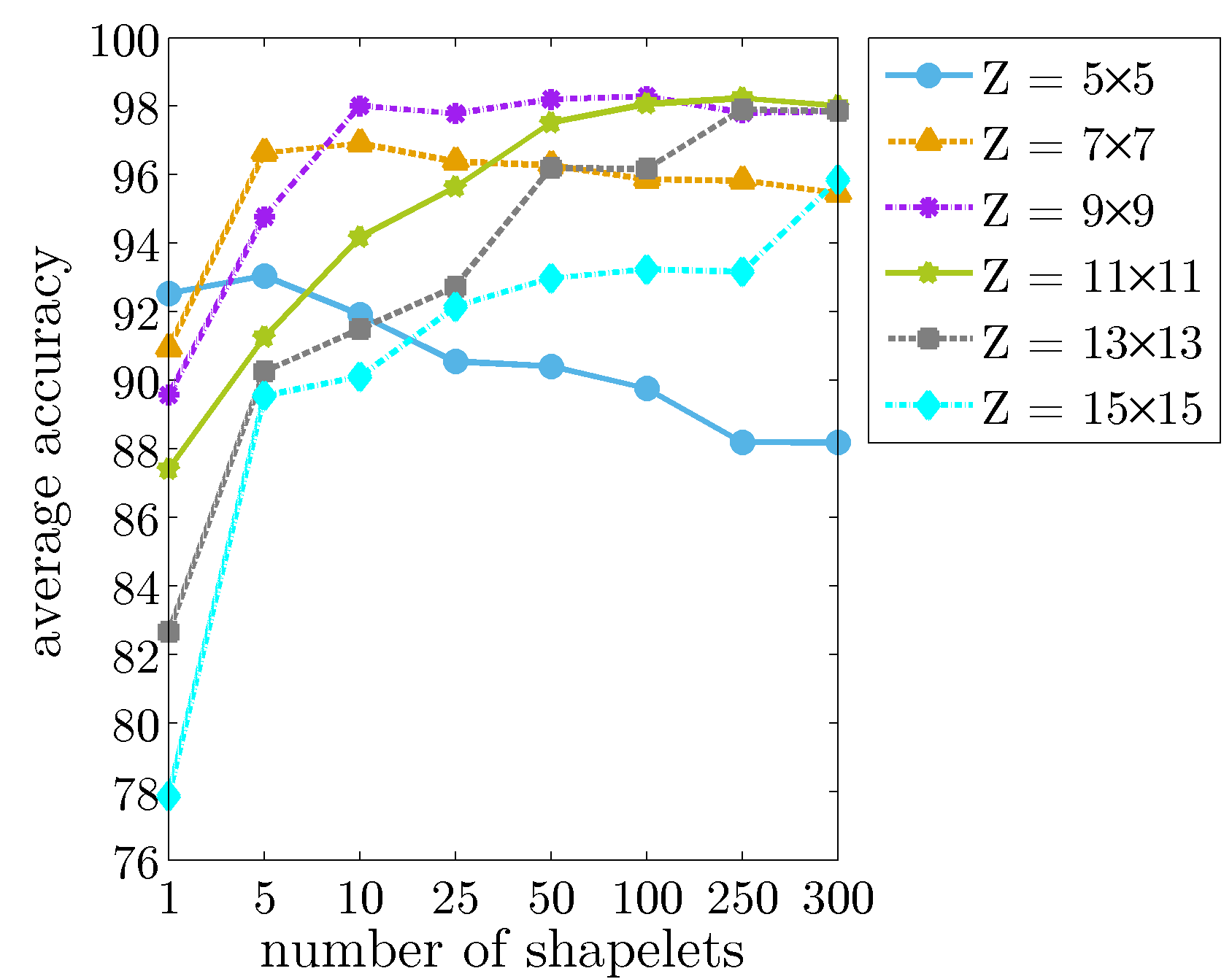}\label{fig:indian_numshape2}}}
	\framebox{\subfigure[Overall accuracy \textsc{University of Pavia}]{\includegraphics[width=0.45\textwidth]{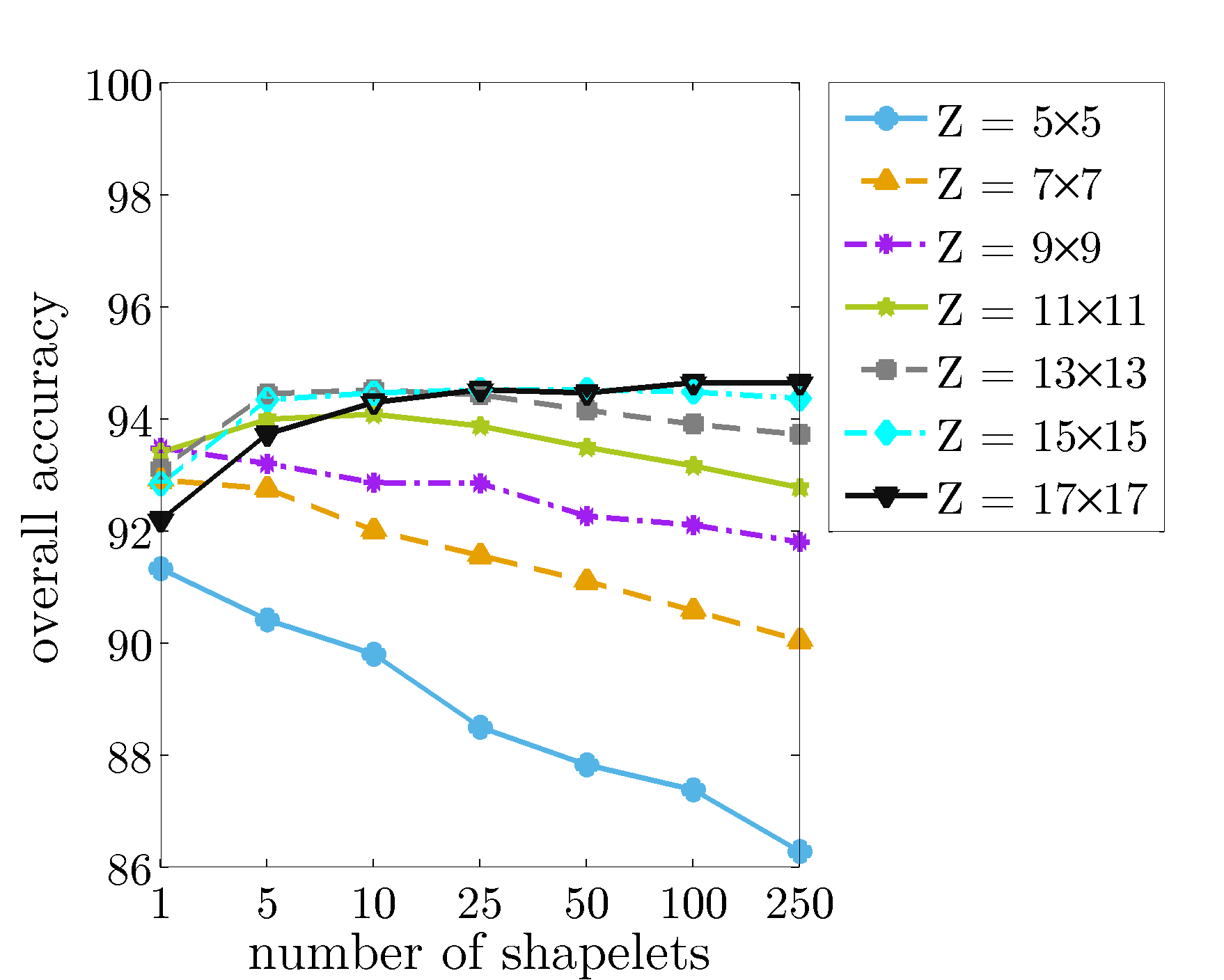}\label{fig:paviaUni_numshape1}}}
	\framebox{\subfigure[Average accuracy \textsc{University of Pavia}]{\includegraphics[width=0.45\textwidth]{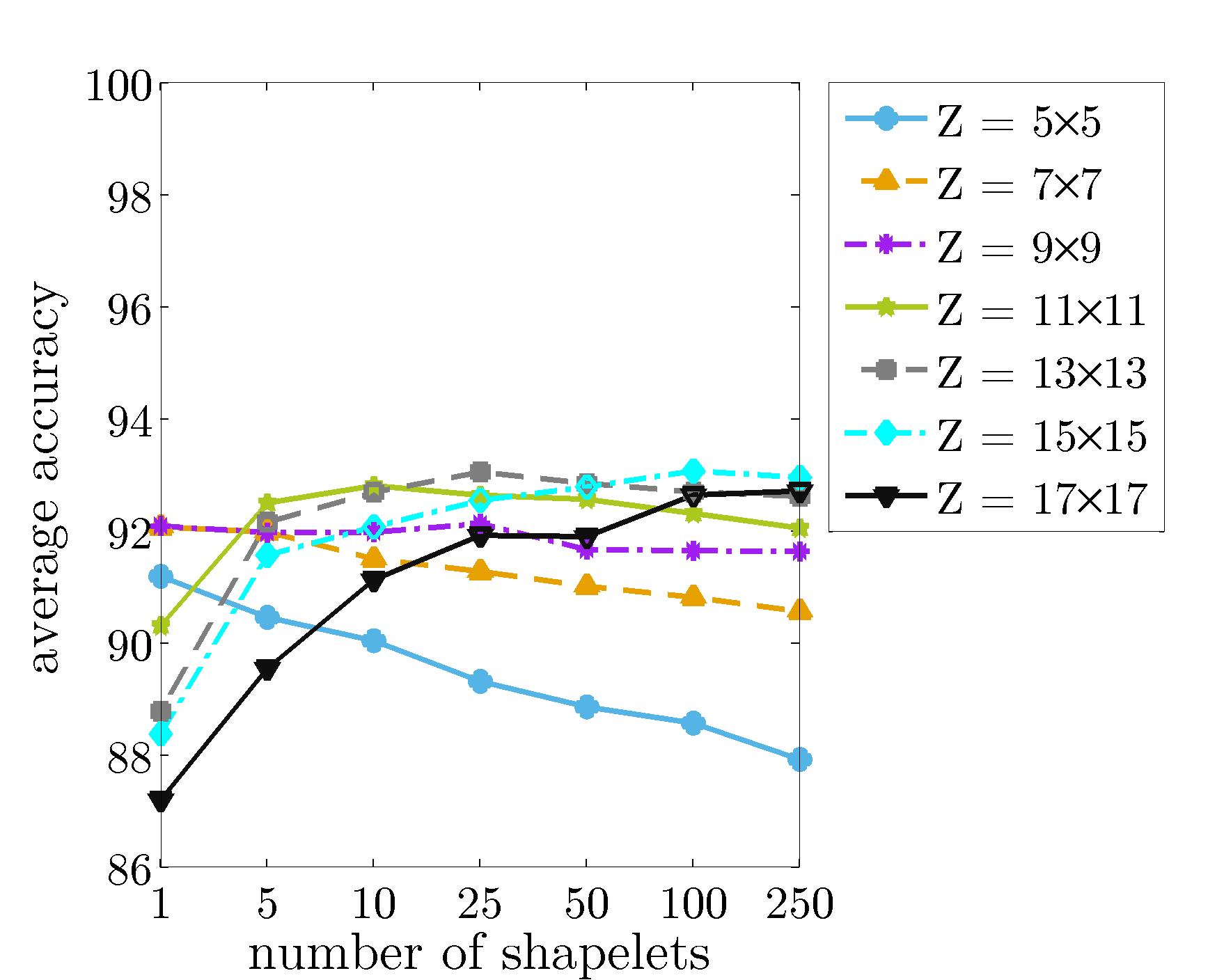}\label{fig:paviaUni_numshape2}}}
	\framebox{\subfigure[Overall accuracy \textsc{Center of Pavia}]{\includegraphics[width=0.45\textwidth]{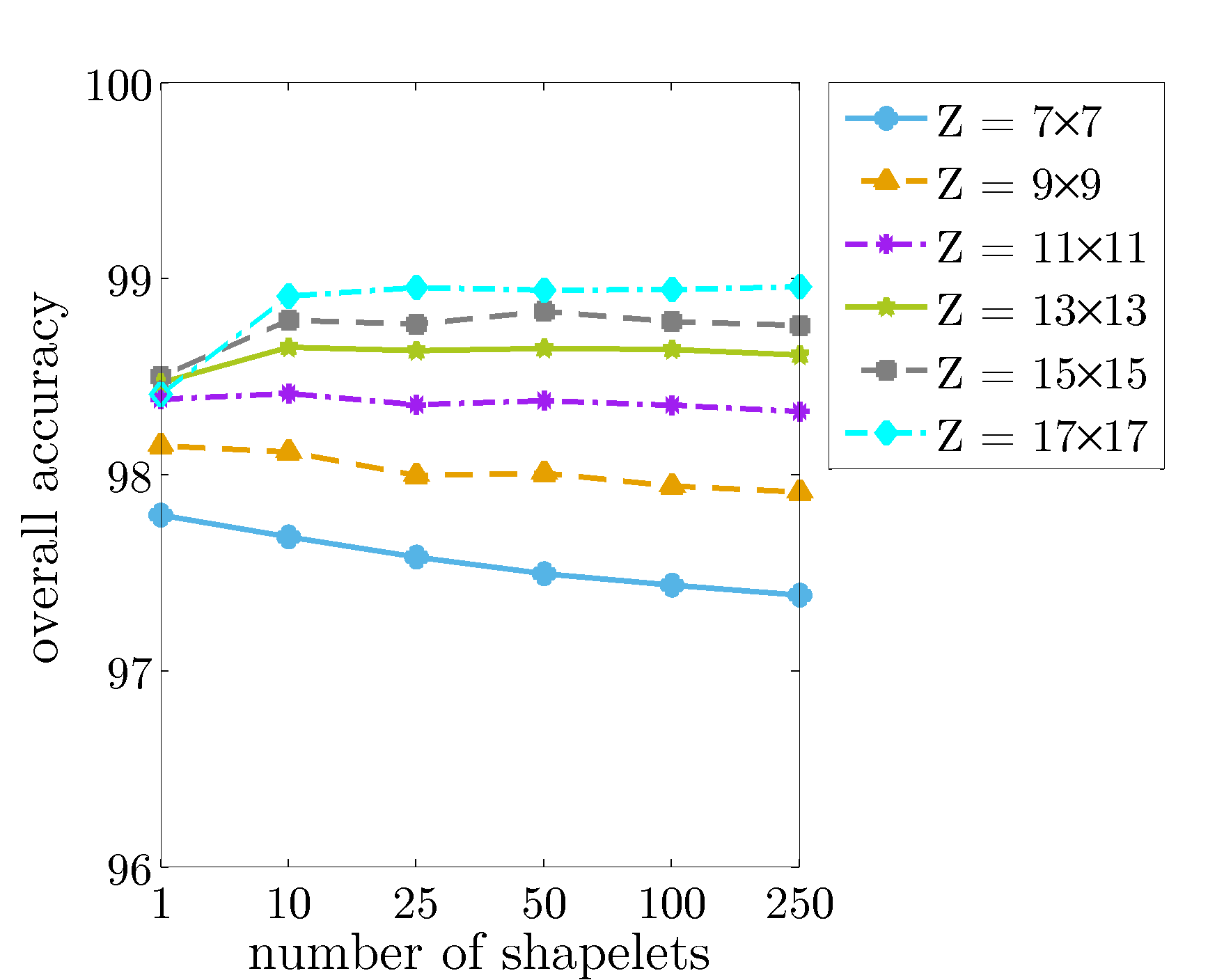}\label{fig:paviaCenter_numshape1}}}	
	\framebox{\subfigure[Average accuracy \textsc{Center of Pavia}]{\includegraphics[width=0.45\textwidth]{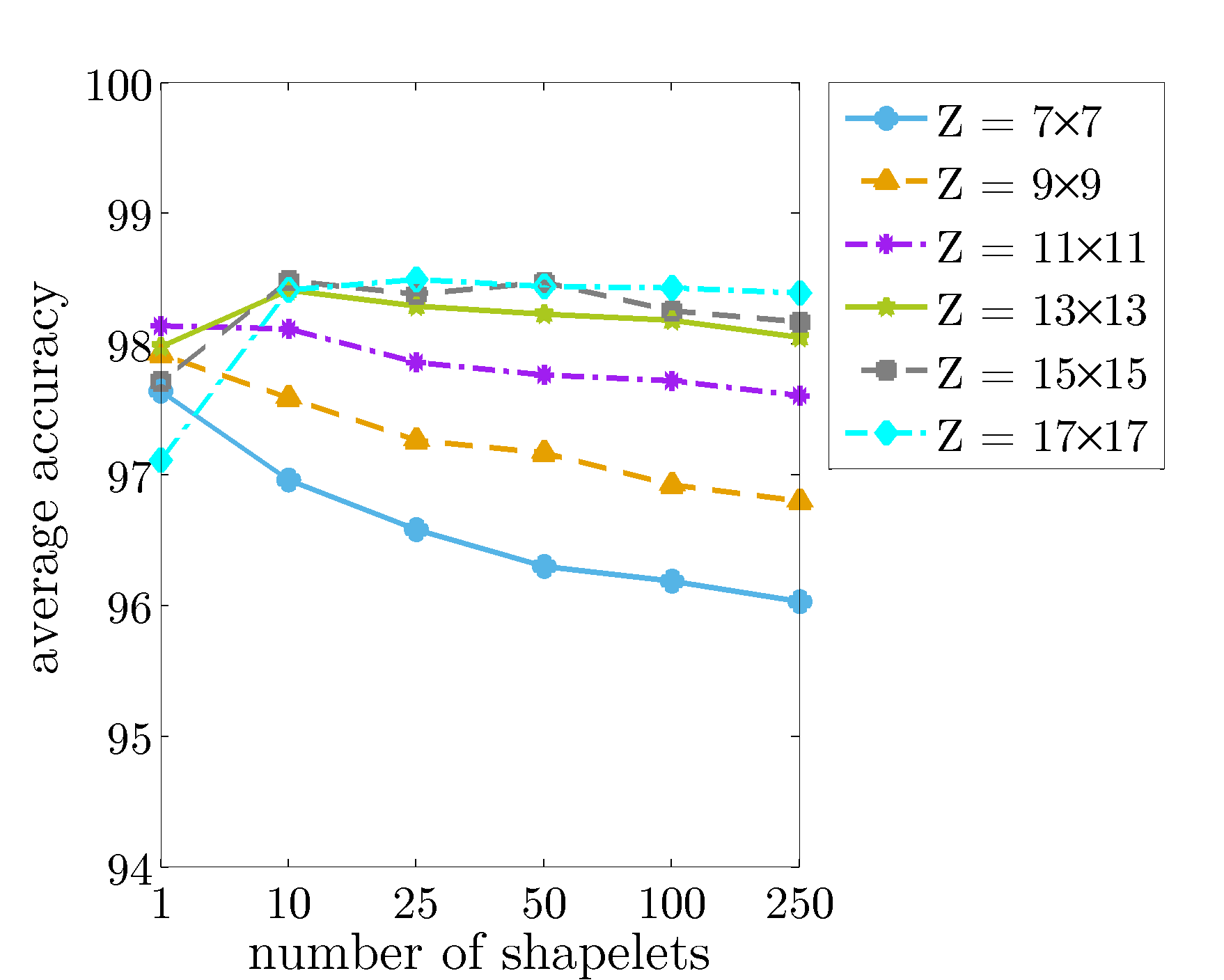}\label{fig:paviaCenter_numshape2}}}
  \caption{Influence of the number of shapelets and the patchsize on the overall accuracy and average accuracy. The number of used dictionary elements is fixed to $W = 3$.}
  \label{fig:numshape}
\end{figure*}

\begin{figure*}[ht]
  \centering
	\framebox{\subfigure[Overall accuracy]{\includegraphics[width=0.45\textwidth]{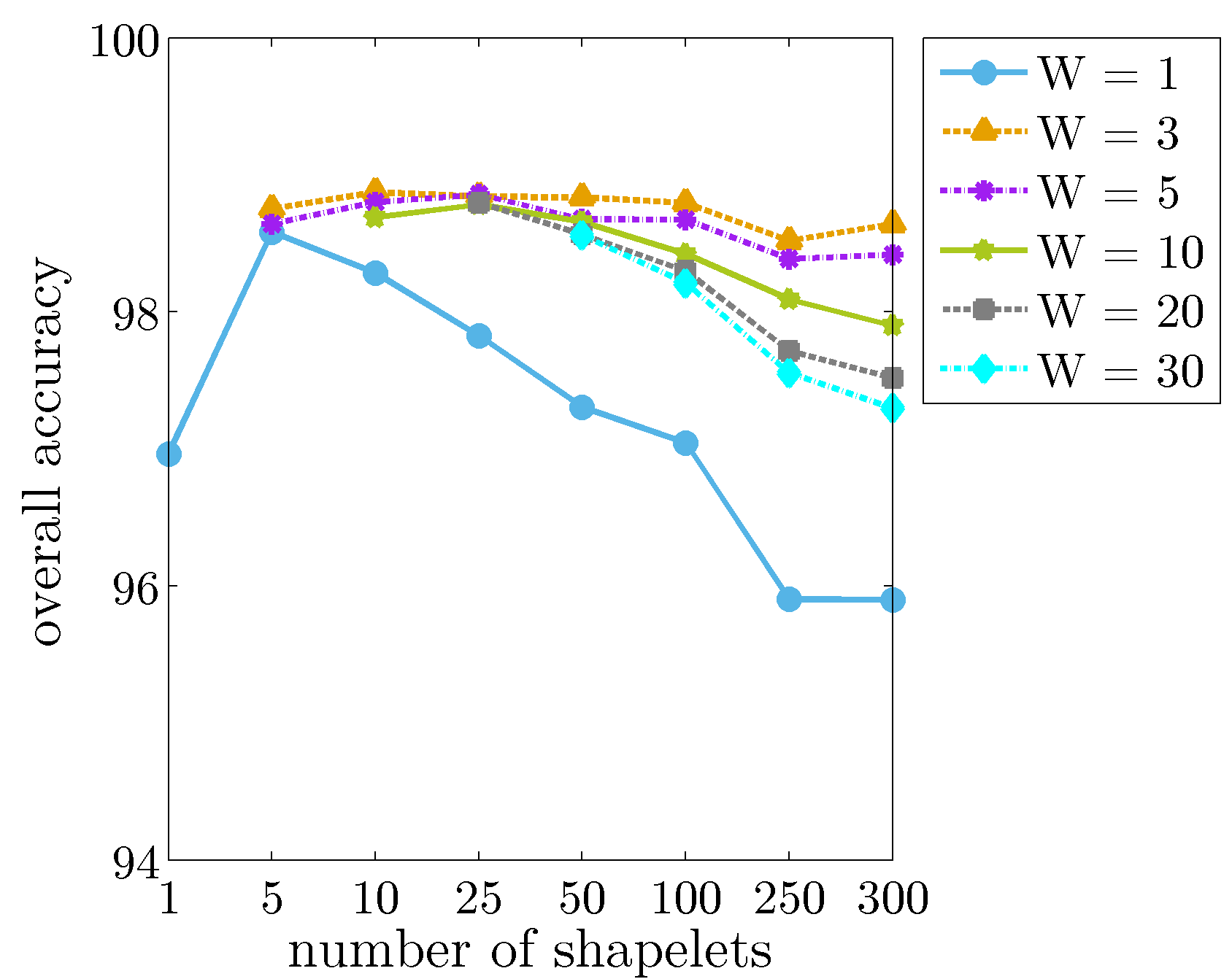}\label{fig:indian_codshape1}}}
	\framebox{\subfigure[Average accuracy]{\includegraphics[width=0.45\textwidth]{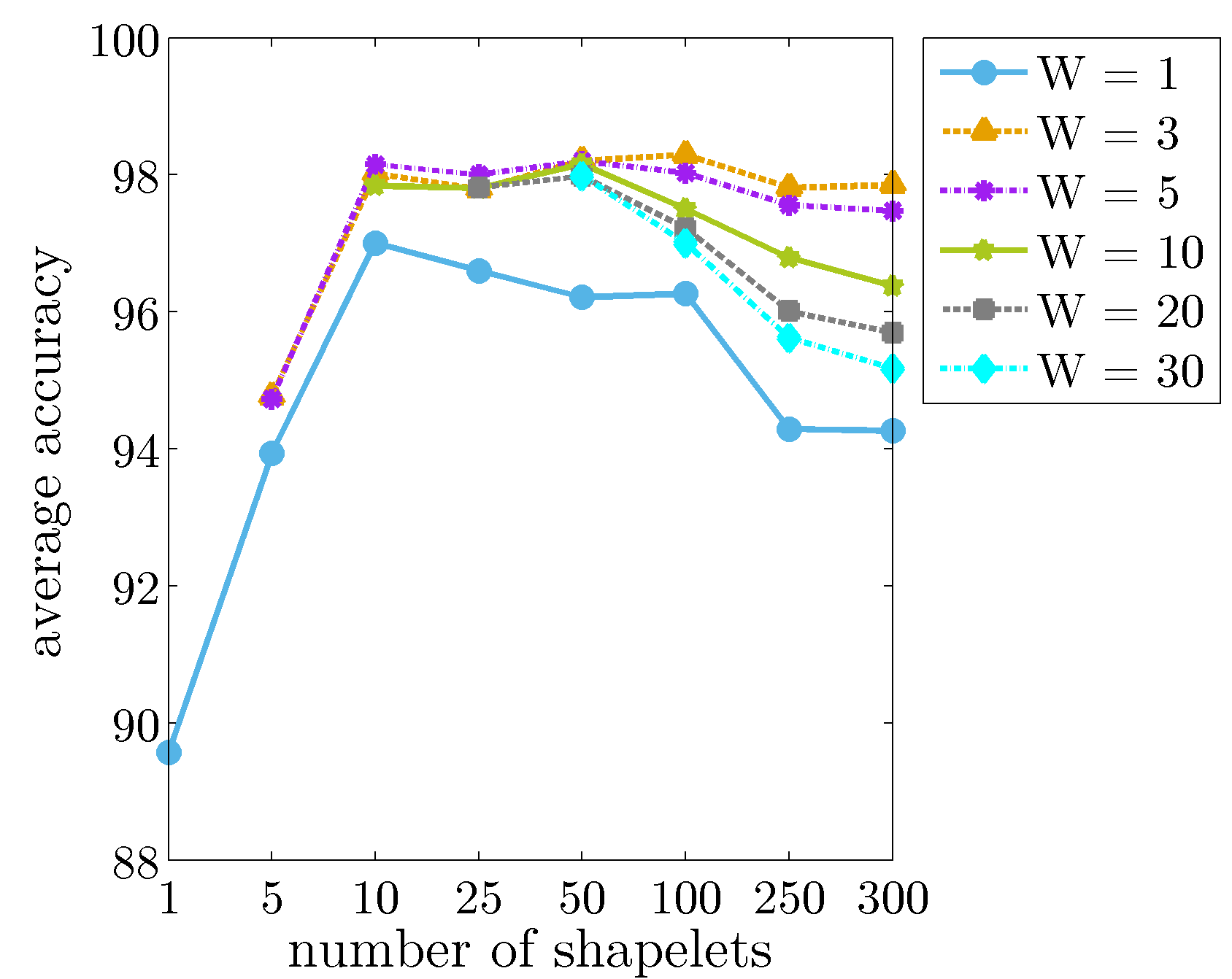}\label{fig:indian_codshape2}}}
  \caption{Impact of the number of used dictionary elements on the overall accuracy and average accuracy for the \textsc{Indian Pines} dataset.}
  \label{fig:numW}
\end{figure*}

Fig. \ref{fig:indian_numshape1} and \ref{fig:indian_numshape2} show the impact of the number of shapelets and the patchsize on the classification accuracies. 
The number of used dictionary elements is fixed to $W = 3$.
The results show that the highest accuracies are obtained with a patchsize of $9\times 9$~pixel.
The best overall accuracy of $99.14\%$ is obtained for a patchsize of $13\times 13$~pixel and 50 shapelets and the highest average accuracy of $98.29\%$ is obtained for a patchsize of $9\times 9$~pixel with 100 shapelets (see Tab. \ref{tab:results_indian}), however, several other parameter settings show comparable results.
The plots clearly indicate the gain of using shapelets for large patchsizes instead of using only one homogeneous patch, \ie, one shapelet.
However, even with only one homogeneous patch our proposed approach can achieve higher accuracies than many of the approaches considered for comparison.
This underlines the fact that it is worth to learn a specifically designed dictionary in order to increase the classification accuracy.
Especially for large patchsizes, which show poor results for a small number of shapelets, the accuracies significantly increase with an increasing number of shapelets.
However, if the patchsize is small an increased number of shapelets results in lower accuracies.
We assume a reason for this to be an overfitting effect, where noise is fitted by potentially non-representative shapelets.
That means the more shapelets are extracted, the less representative they are for the image.
The effect becomes the less apparent the larger the patchsize is, since the number of used dictionary elements is restricted to $W=3$ and due to this, large patches are hardly able to model noise.
This is underlined by the results presented in Fig. \ref{fig:numW}.
The patchsize is fixed to $9 \times 9$, while the influence of the number of dictionary elements is evaluated by means of the overall and average accuracy.
The highest accuracy is achieved with 3 dictionary elements (\ie, $W = 3$), while the accuracy decreases significantly when the number of elements is further increased.

\begin{figure*}[ht]
  \centering
	\framebox{\subfigure[Overall accuracy]{\includegraphics[width=0.45\textwidth]{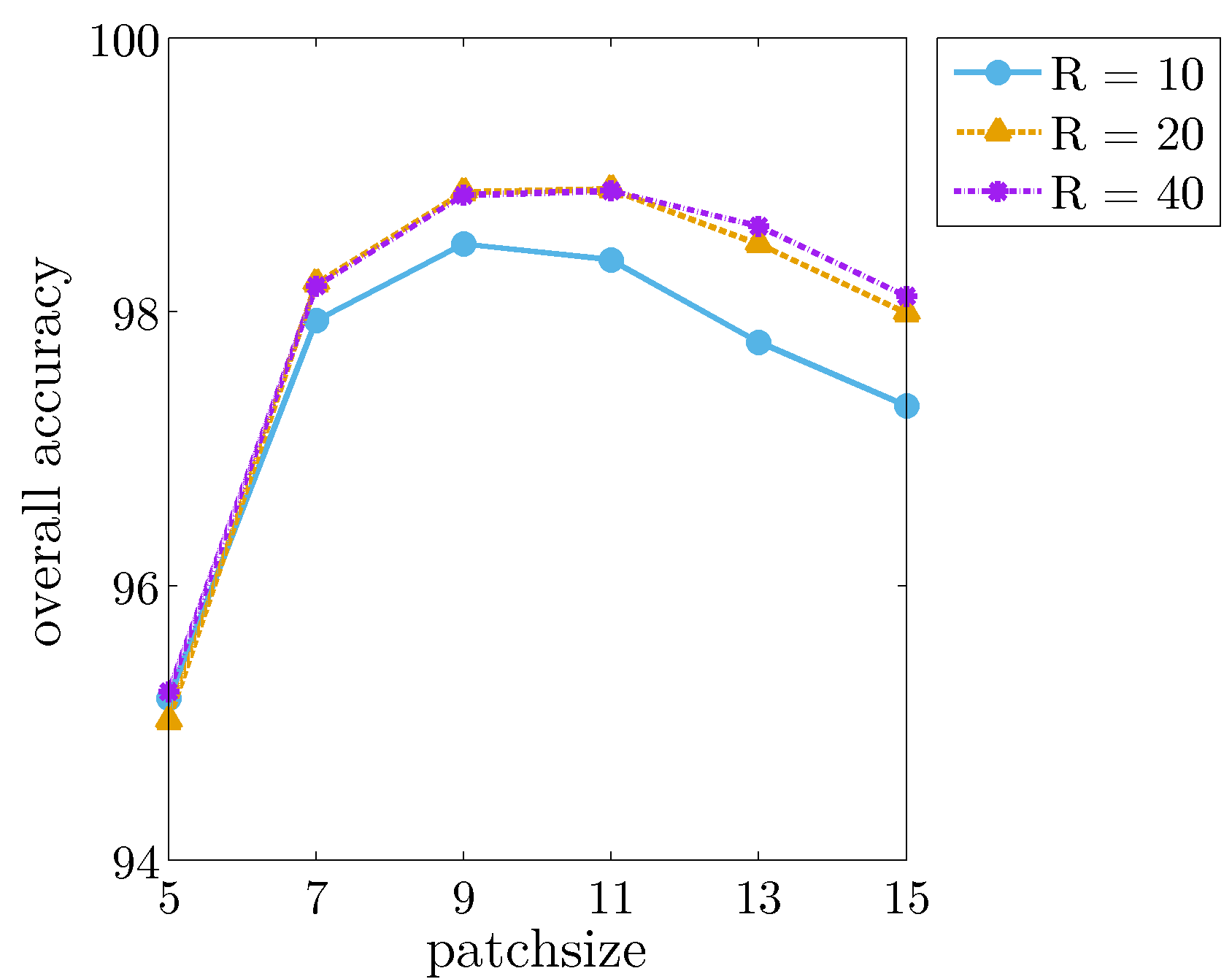}}}
	\framebox{\subfigure[Average accuracy]{\includegraphics[width=0.45\textwidth]{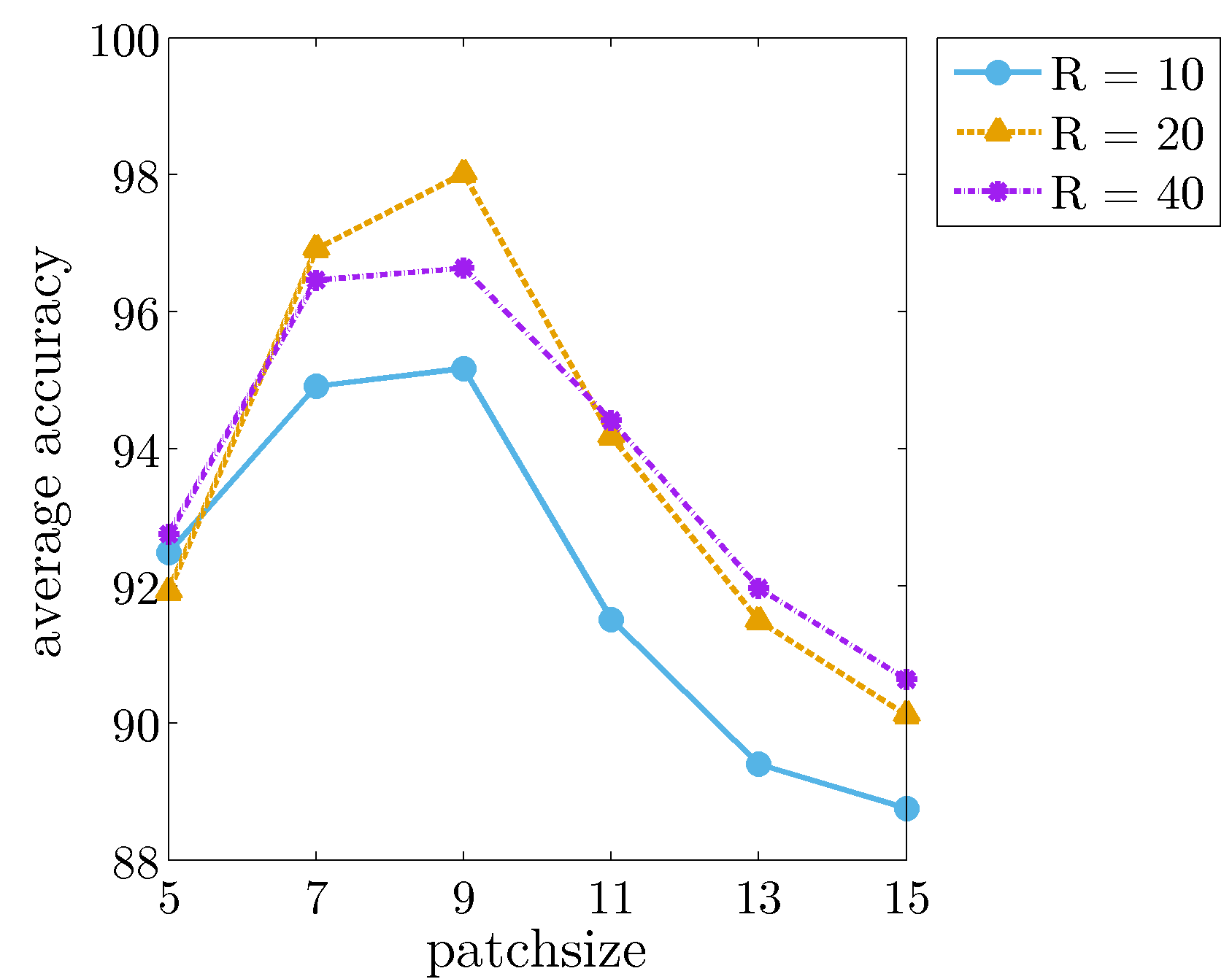}}}
  \caption{Impact of approximate superpixel size for shapelet extraction on the overall accuracy and average accuracy for the \textsc{Indian Pines} dataset.}
  \label{fig:SPsize}
\end{figure*}

Fig. \ref{fig:SPsize} shows the impact of the approximate superpixel size $R$ for shapelet extraction on the classification accuracies, while the number of shapelets is fixed to $10$.
An approximate size of $20\times 20$~pixel shows good results for all patchsizes, whereas an approximate size of $10 \times 10$~pixel provide less representative shapelets. 
A larger size of approximately $40\times 40$~pixel works well for larger patchsizes indicating that the approximate superpixel size should be increased with an increased patchsize.

\subsection{\textsc{University of Pavia} Dataset}
Tab. \ref{tab:results_paviaUni} shows the result obtained by using $100$ shapelets and a patchsize of $15\times 15$ pixels.
Our proposed approach clearly outperforms the other methods in terms of overall accuracy and $\kappa$ coefficient (see Tab. \ref{tab:results_paviaUni} and Fig. \ref{fig:results}). 
This is also underlined by the class-specific accuracies. 
In many cases the approach performs better or at least equally well when compared to the class accuracies achieved by the other methods. 
As for the \textsc{Indian Pines} dataset the results clearly show that the use of shapelets significantly improves the classification accuracies for large patchsizes when compared to similar approaches assuming only a homogeneous neighborhood (\eg, SOMP). 
The most accurate results are achieved by a relatively large patchsize of $15\times 15$~pixel (see Fig. \ref{fig:paviaUni_numshape1} and Fig. \ref{fig:paviaUni_numshape2}) and thus, the classification result is very smooth within large homogeneous regions (\eg, \textsc{meadows}).
Nevertheless, also small regions such as \textsc{shadows} and \textsc{metal sheets} are classified with high accuracies. 
The main reason for this might be the usage of shapelets.
Besides our proposed approach also SCDL uses a large patchsize of $16 \times 16$~pixel, while all other approaches (SVMCK, SOMP, SSP and kernel simultaneous subspace pursuit (KSSP)) use a smaller patchsize of $5\times 5$~pixel (\cite{Soltani-Farani2013,Chen2013}).
As already stated by \cite{Chen2011}, unlike the \textsc{Indian Pines} dataset this dataset mostly lacks in large spatial homogeneous regions and thus, a small patchsize is used for SVMCK, SOMP, SSP and KSSP in order to represent small regions. 
Although the number of shapelets producing the best result is relatively high, the actual size of the dictionary is lower since some spatial-spectral dictionary elements are redundant and can be removed.
For the best result, the average number of dictionary elements is $45$ with a standard deviation of $33$ elements. 

\begin{table*}[ht]
\tabcolsep2pt
\centering
\footnotesize
\caption{Size of training and test data, classwise accuracies [\%], overall accuracy [\%], average accuracy [\%] and $\kappa$ coefficient of \textsc{University of Pavia} dataset. Bold number indicates the best results and underlined results the second best result.}
\begin{tabular}{lcccccccccc}
	& \# train & \# test & SVMCK & SOMP & SSP & KSPCK & KSSP & KOMPCK & SCDL & \name\\
		\hline
\colorbox{pu1}{\textcolor{pu1}{A}} Asphalt & 548 & 6304 	& 79.85 & 59.33 & 69.59 & 89.64 & \underline{89.56} & 82.23 & 81.87 & \textbf{96.64}\\
\colorbox{pu2}{\textcolor{pu2}{A}} Meadows & 540 & 18146 & 84.86 	& 78.15 & 72.31 & 72.68 & 79.98 & 72.47 & \underline{96.48} & \textbf{98.50} \\
\colorbox{pu3}{\textcolor{pu3}{A}} Gravel & 392 & 1815 	 & 81.87 & \underline{83.53}  & 74.10 & 80.06 &  \textbf{85.45} & 82.26 &  83.36 & 76.31\\
\colorbox{pu4}{\textcolor{pu4}{A}} Trees & 524 & 2912 & 96.36 & 96.91  & 95.33 & \textbf{98.94} &  \underline{98.66} & 98.56 & 95.47 & 91.28 \\
\colorbox{pu5}{\textcolor{pu5}{A}} Metal sheets 	& 265 & 1113 & 99.37 & 99.46  & 99.73 & \textbf{100.00} &  \underline{99.91} & 99.82 &  99.82 & \textbf{100.00}\\
\colorbox{pu6}{\textcolor{pu6}{A}} Bare soil & 532 & 4572 & 93.55 & 77.41 & 86.72 & \underline{94.77} &  \textbf{95.76} & 93.92 &  81.21 & 83.81\\
\colorbox{pu7}{\textcolor{pu7}{A}} Bitumen & 375 & 981 & 90.21 & \underline{98.57} & 90.32 & 89.81 &  97.96 & 92.46 &  74.11 & \textbf{100.00} \\
\colorbox{pu8}{\textcolor{pu8}{A}} Bricks & 514 & 3364 	& 92.81 & 89.09 & 90.46 & 89.54 &  \textbf{96.43} & 78.78 &  85.91 & \underline{95.33}\\
\colorbox{pu9}{\textcolor{pu9}{A}} Shadows & 231 & 795 & 95.35 & 91.95 & 90.94 & 96.48  &  \textbf{98.49} & 96.98 &  96.60 & \underline{97.74} \\
\hline
Overall 			& 3921 & 40002 	& 87.18 & 79.00 & 78.39 & 83.19 &  87.65 & 81.07 &  \underline{90.42} & \textbf{94.48}\\
Average 	    		&  & 			& 90.47 &  86.04 & 85.50 & 90.21 &  \textbf{93.58} & 88.61 &  88.31 & \underline{93.07} \\
$\kappa$ 		& & 				& 0.833 & 0.728 & 0.724 & 0.785 &  0.840 & 0.785 &  \underline{0.870} & \textbf{0.925}\\
\end{tabular}
\label{tab:results_paviaUni}
\end{table*}
    
In order to analyze the influence of the quality of the segmentation, we replaced the image-specific shapelets with Haar wavelets as presented by \cite{Aharon2006}, which usually do not represent characteristic patterns in the image.
The overall accuracy decreases to $93.62$, the average accuracy to $88.84$ and $\kappa$ to 0.912.
However, the results are still better or comparable to the presented approaches. 
To our experience every segmentation algorithm which can roughly represent the boundaries between different classes within the image leads to an increase in accuracy over results obtained by using synthetic shapelets such as Haar wavelets.
    
\subsection{\textsc{Center of Pavia} Dataset}
\begin{figure*}[ht]
  \centering
  	\framebox{\subfigure[Image]{\includegraphics[width=0.18\textwidth]{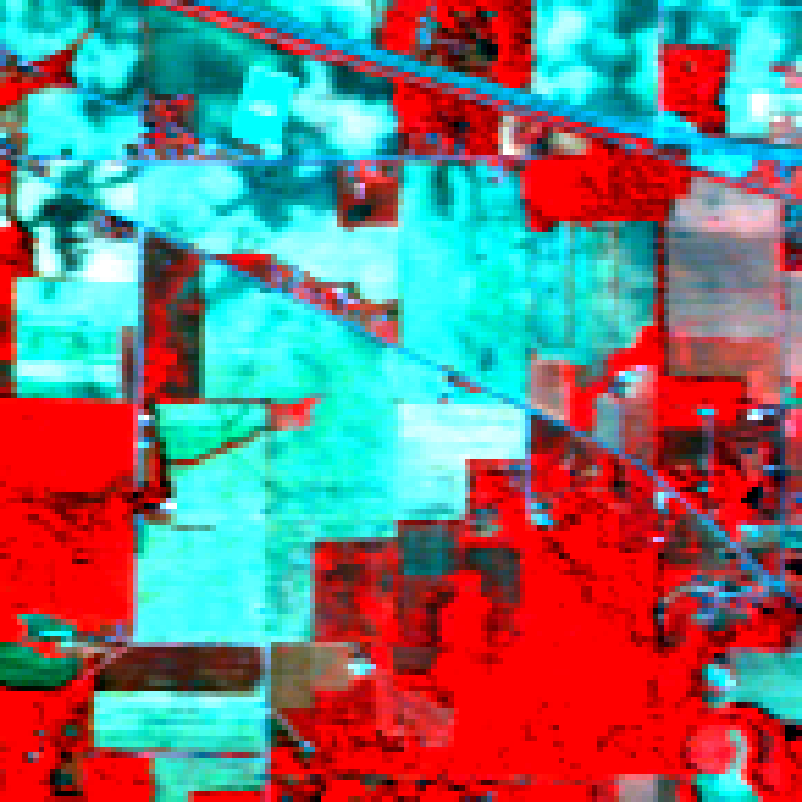}}}
	\framebox{\subfigure[Training data]{\includegraphics[width=0.18\textwidth]{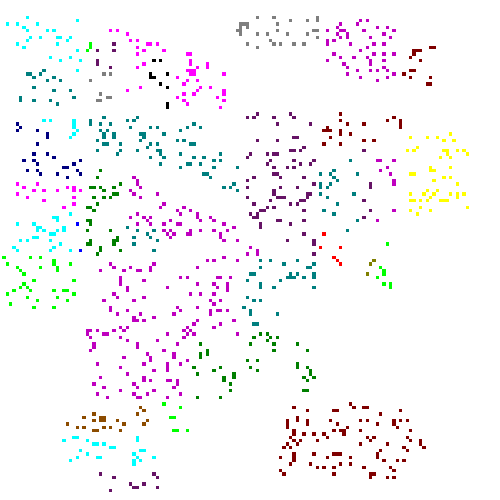}}}
	\framebox{\subfigure[Test data]{\includegraphics[width=0.18\textwidth]{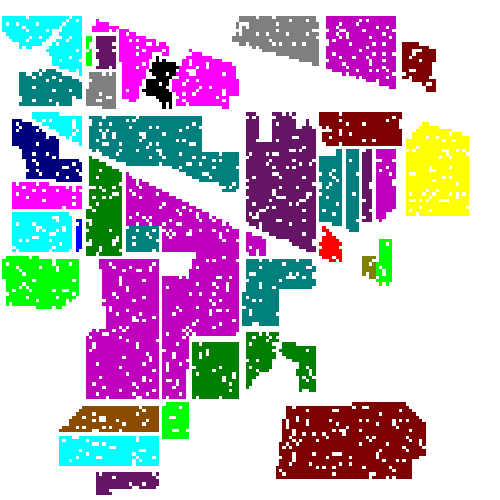}}}
	\framebox{\subfigure[\name~result]{\includegraphics[width=0.18\textwidth]{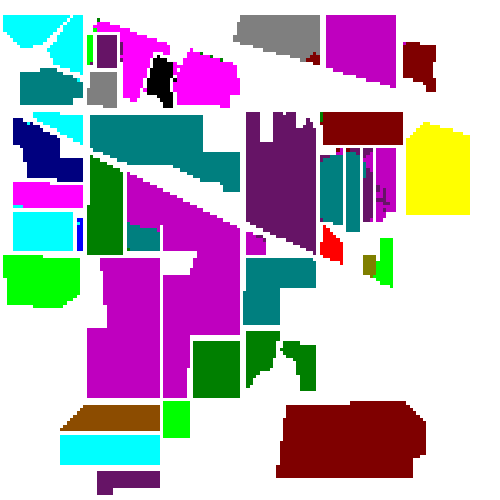}}} 
	
	\framebox{\subfigure[Image]{\includegraphics[width=0.18\textwidth]{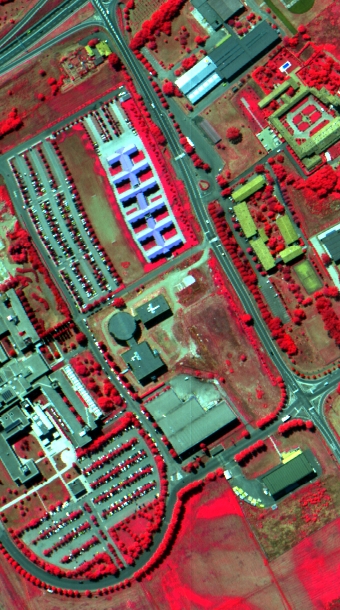}}}
	\framebox{\subfigure[Training data]{\includegraphics[width=0.18\textwidth]{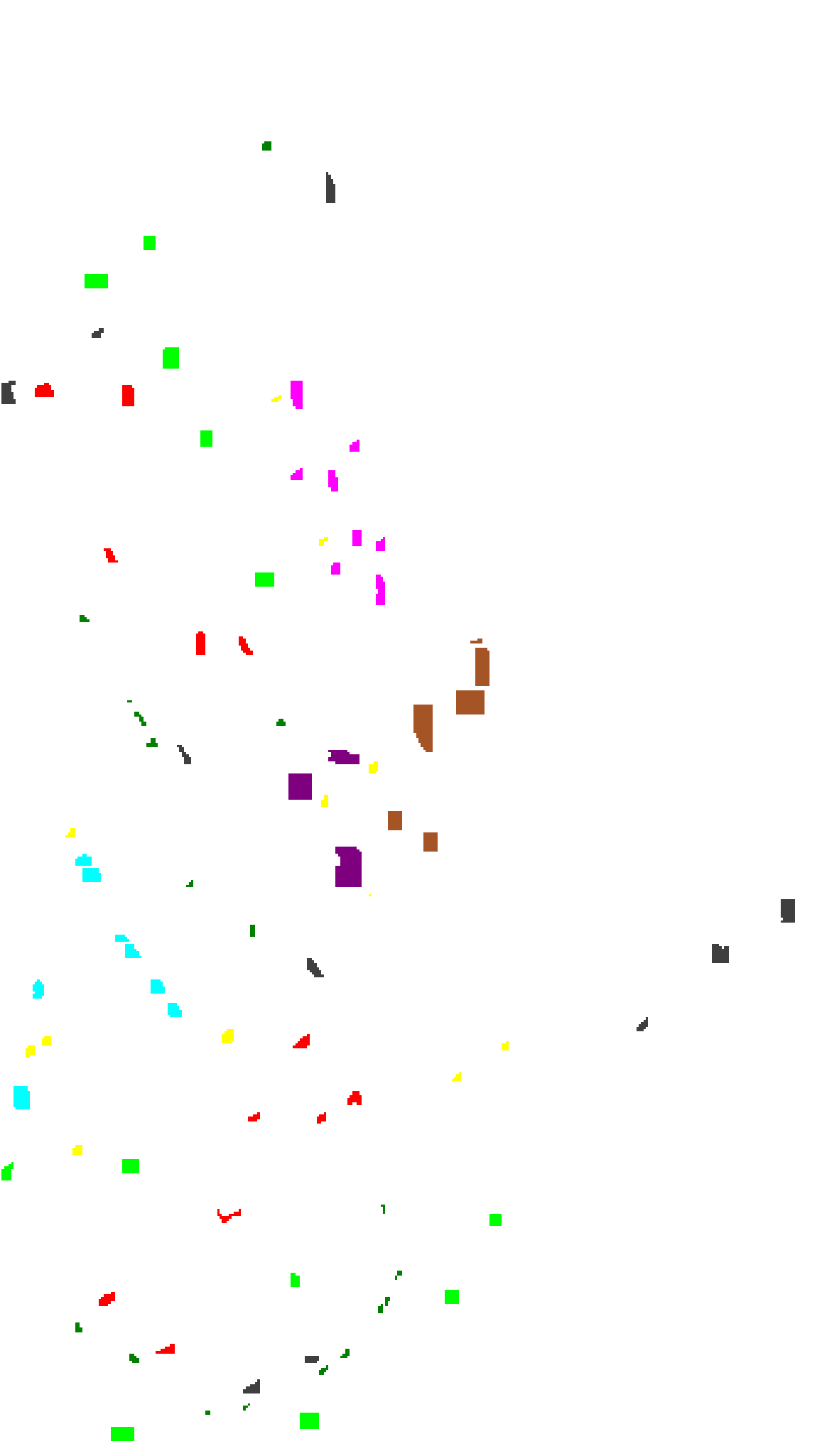}}}
	\framebox{\subfigure[Test data]{\includegraphics[width=0.18\textwidth]{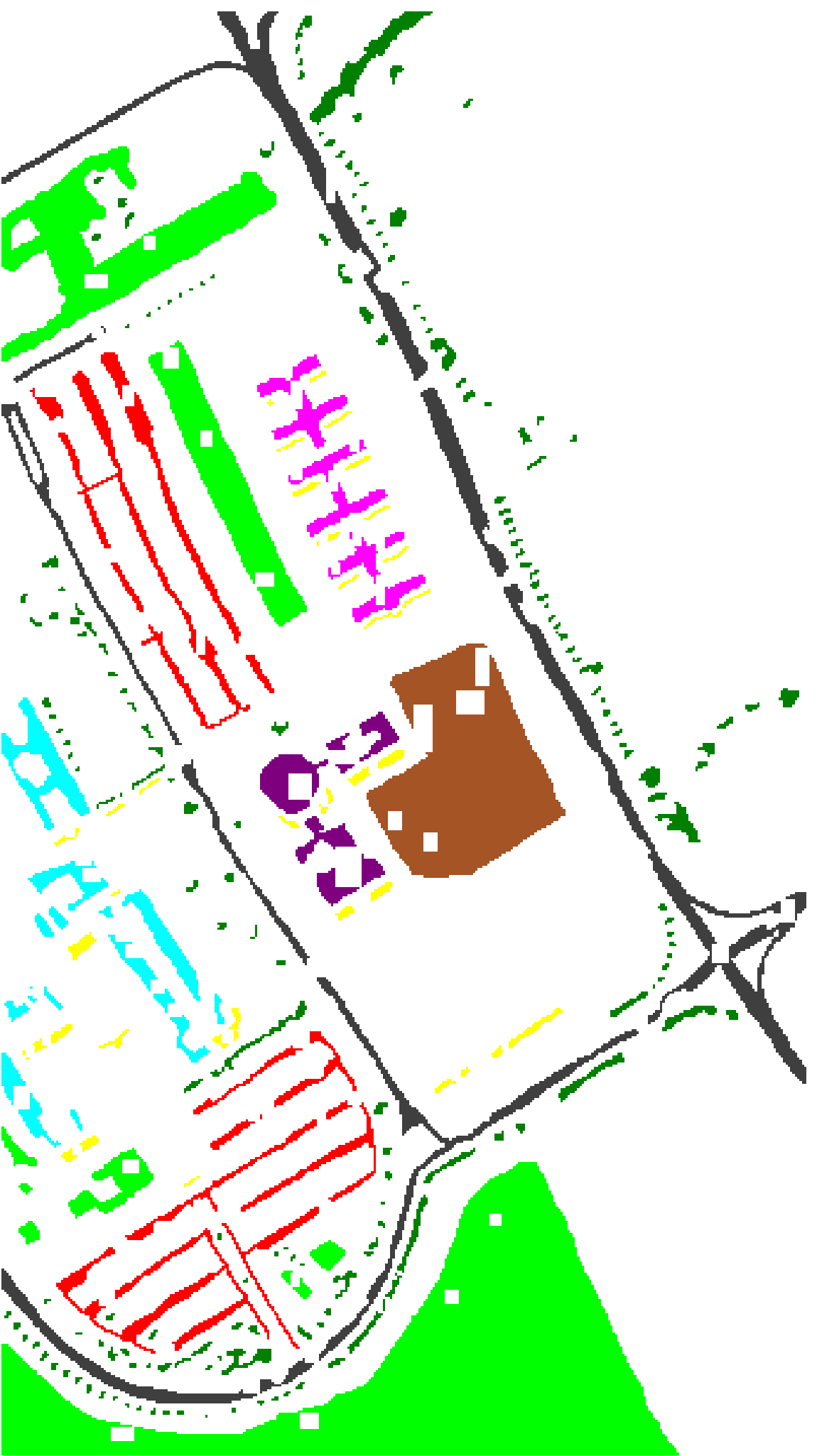}}} 
	\framebox{\subfigure[\name~result]{\includegraphics[width=0.18\textwidth]{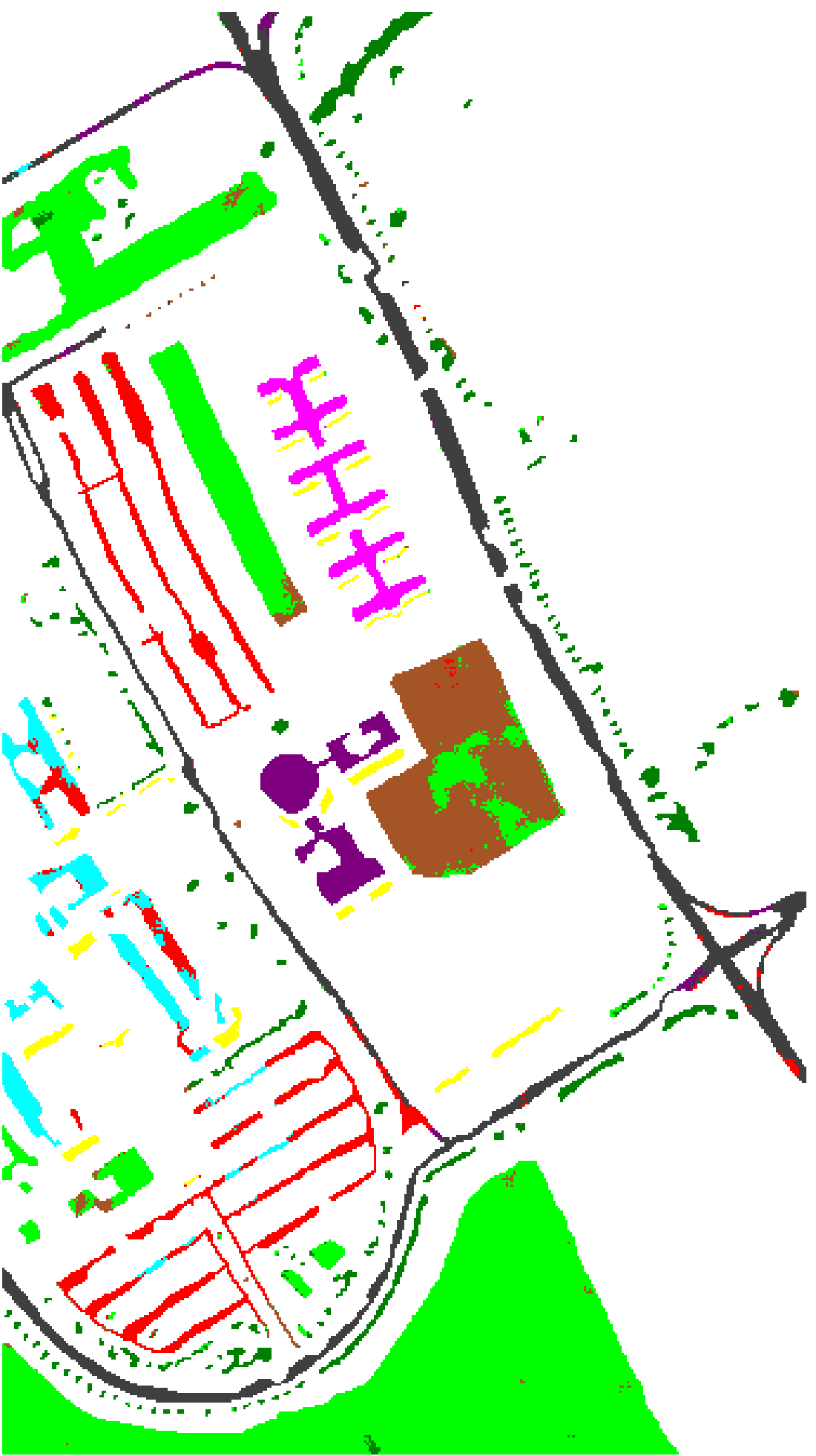}}} 
	
	\framebox{\subfigure[Image]{\includegraphics[width=0.18\textwidth]{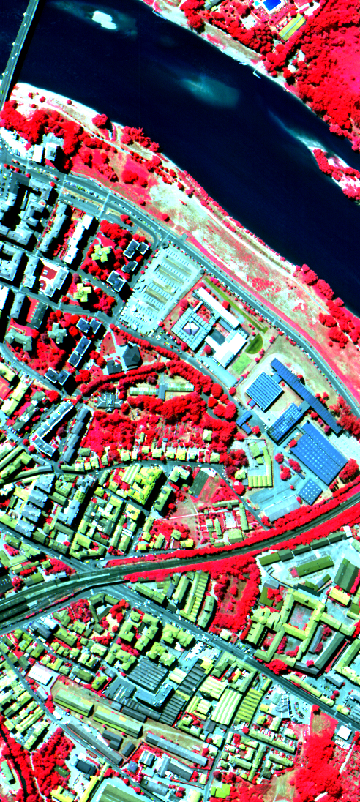}}}
	\framebox{\subfigure[Training data]{\includegraphics[width=0.18\textwidth]{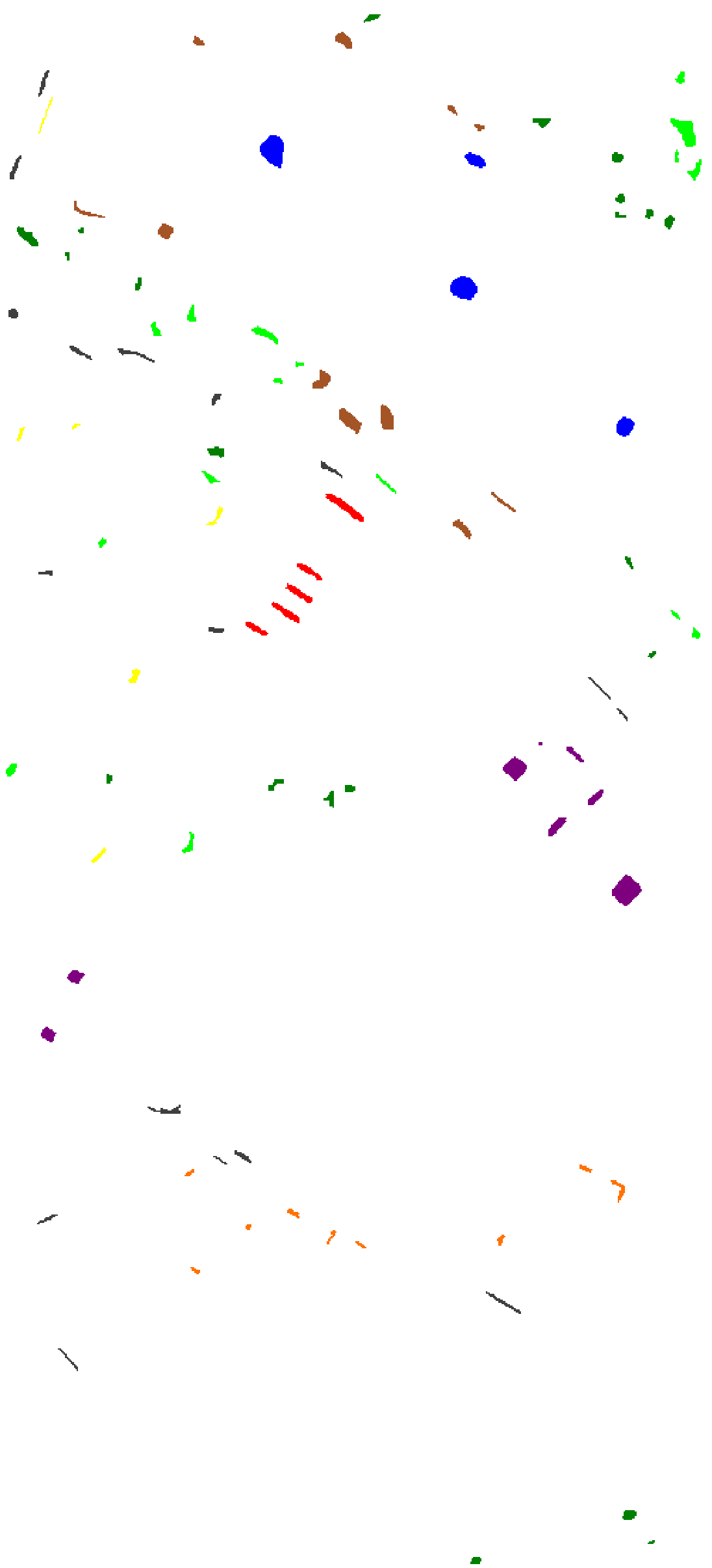}}}
	\framebox{\subfigure[Test data]{\includegraphics[width=0.18\textwidth]{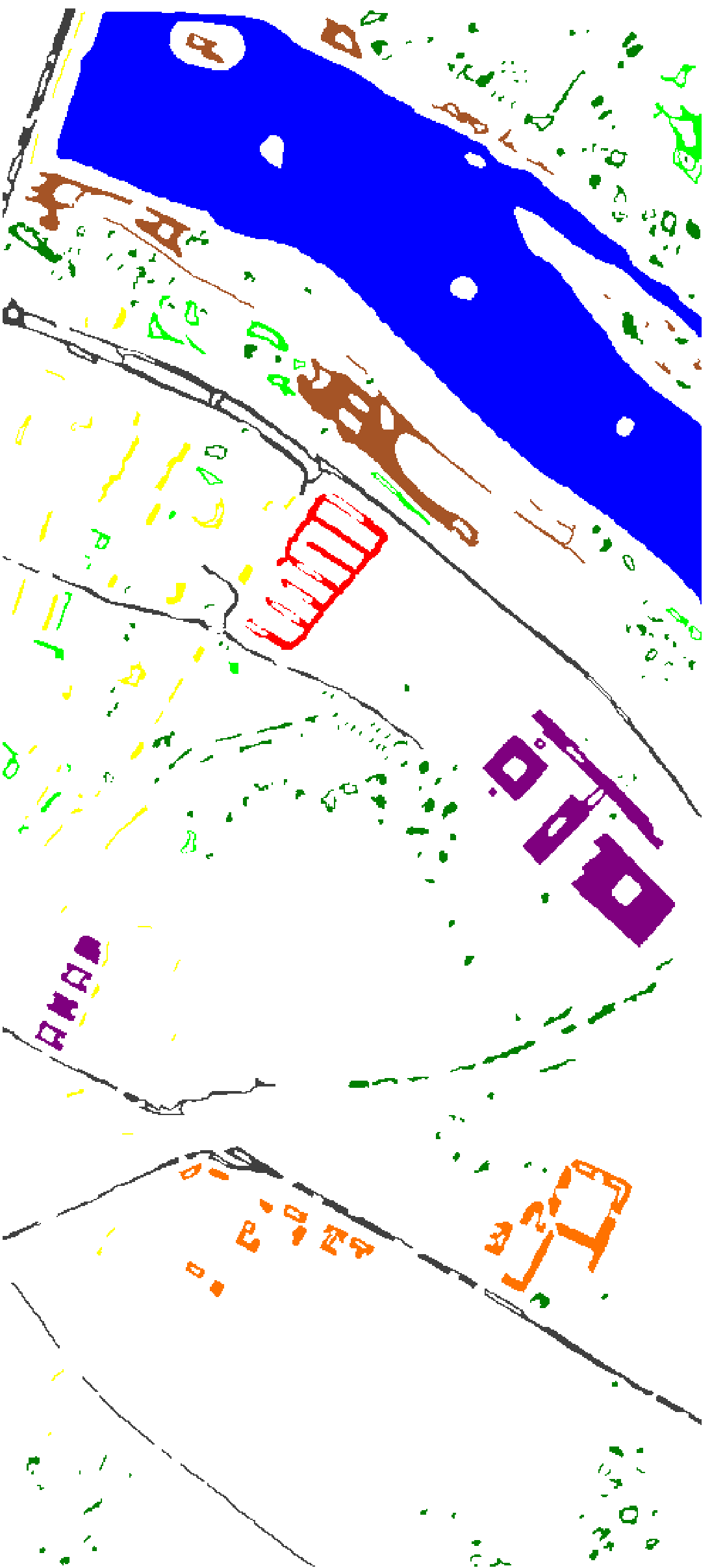}}}
	\framebox{\subfigure[\name~result]{\includegraphics[width=0.18\textwidth]{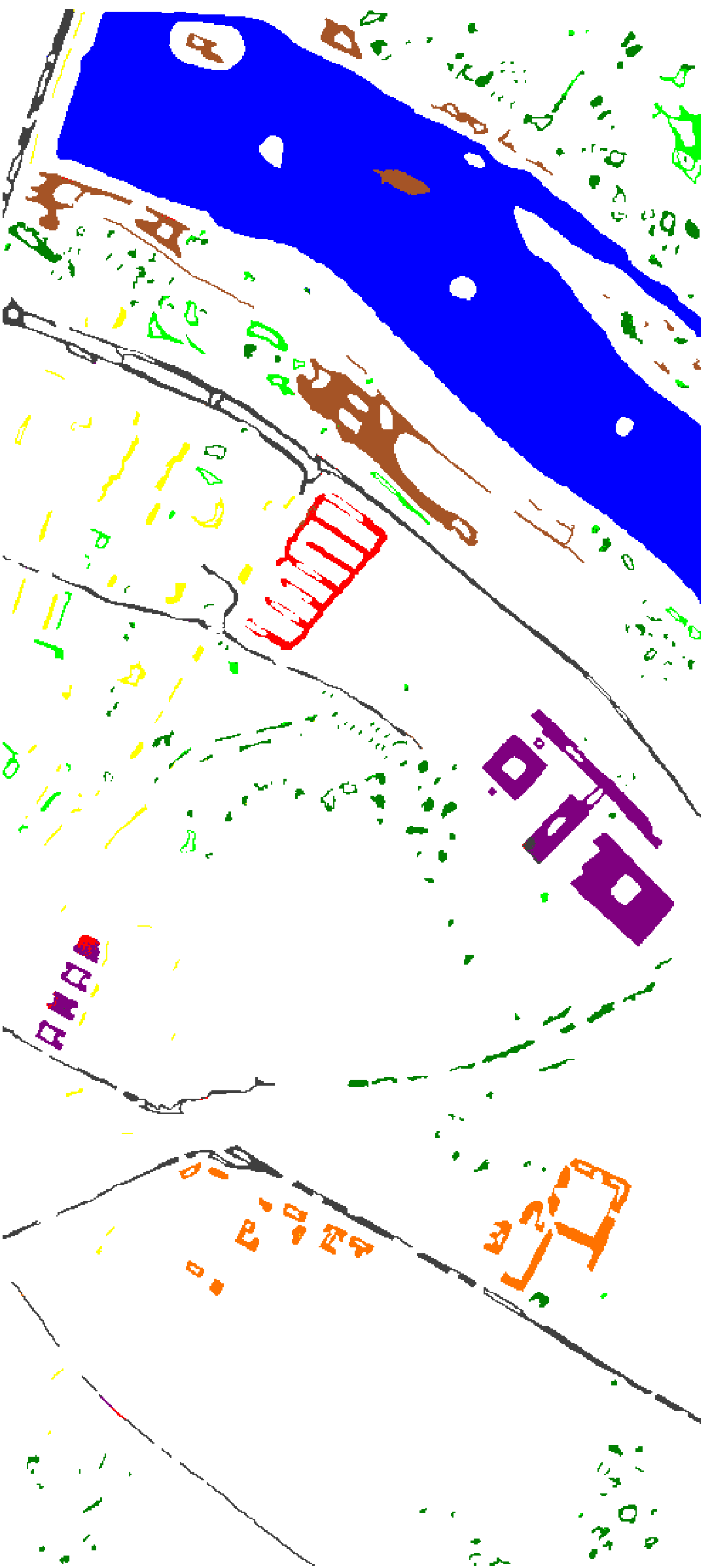}}}		
  \caption{Image, training and test data of the datasets \textsc{Indian Pines} dataset (upper row), \textsc{University of Pavia} (middle row) and \textsc{Center of Pavia} (bottom row). The classification result is presented in the right column.}
  \label{fig:results}
\end{figure*}

Tab. \ref{tab:results_paviaCenter} shows the result obtained by using $25$ shapelets and a patchsize of $17\times 17$ pixels.
As before, our approach performs better than compared methods in terms of overall accuracy, average accuracy and $\kappa$ or performs at least equally well. 
This is also underlined by the class-specific results achieved by our proposed method. 
In the very most cases the highest or at least the second-best class accuracy is achieved, resulting in more balanced results.
As for the other results, Fig. \ref{fig:results} shows that classes with large homogeneous regions (\eg, \textsc{water}) as well as small classes (\eg, \textsc{asphalt}) can be classified accurately.

Besides the before mentioned results, as illustrated in Fig. \ref{fig:paviaCenter_numshape1}, the highest overall accuracy could be obtained with $250$ shapelets using a patchsize of $17\times 17$~pixel.
Although the number of shapelets in this case is relatively high, the average number of spatial-spectral dictionary elements after removing redundant elements is $45$ with a standard deviation of $69$ elements. 
As for all dataset, this is significantly lower than for all compared methods, which use mostly the complete training data set as dictionary or at least several hundreds (\eg, SCDL).

As for \textsc{University of Pavia} dataset, we analyzed the influence of the quality of segmentation by replacing the shapelets by Haar wavelets.
The overall accuracy reduces to $97.58$, the average accuracy to $97.29$ and $\kappa$ to $0.954$.
The obtained results are comparable to most of the approaches.
These results underline the gain in using image-specific shapelets rather than synthetic ones.

\begin{table*}[ht]
\tabcolsep2pt
\centering
\scriptsize
\caption{Size of training and test data, classwise accuracies [\%], overall accuracy [\%], average accuracy [\%] and $\kappa$ coefficient of \textsc{Center of Pavia} dataset. Bold number indicates the best results and underlined results the second best result.}
\begin{tabular}{lccccccccccc}
	& \# train & \# test & SVMCK & SOMP & SSP & KSPCK & KSSP & KOMPCK & SCDL & \name\\
		\hline
\colorbox{pc1}{\textcolor{pc1}{A}} Water & 745 & 64533 & 97.46 & \underline{99.32} & 97.79 & 98.79 & 99.26 & 98.98 & \textbf{99.40} & 99.11\\
\colorbox{pc2}{\textcolor{pc2}{A}} Trees& 785 & 5722 & 93.08 & 92.38 & 92.82 & 91.70 & 91.23 & \textbf{96.31} & 93.19 & \underline{94.73}\\
\colorbox{pc3}{\textcolor{pc3}{A}} Meadow & 797 & 2094 & 97.09 & 95.46  & \underline{97.80} & \textbf{99.57} & 97.71 & 96.08 & 97.49 & 97.43\\
\colorbox{pc4}{\textcolor{pc4}{A}} Brick & 485 & 1667 & 77.02 & 85.66  & 78.52 & 94.54 & 95.26 & 97.78 & \underline{98.32} & \textbf{98.61}\\
\colorbox{pc5}{\textcolor{pc5}{A}} Soil & 820 & 5729 & 98.39 & 96.37  & 95.81 & 94.99 & 97.45 & 97.82 & \underline{99.27} & \textbf{99.98}\\
\colorbox{pc6}{\textcolor{pc6}{A}} Asphalt & 678 & 6847 & 94.32 & 93.81 & 96.52 & 93.92 & \underline{97.41} & 96.54 & 95.45 & \textbf{99.22}\\
\colorbox{pc7}{\textcolor{pc7}{A}} Bitumen & 808 & 6479 & 97.50 & 94.68 & 95.96 & 96.90 & \underline{97.82} & \textbf{98.63} & 95.77 & 97.07\\
\colorbox{pc8}{\textcolor{pc8}{A}} Tile & 223 & 2899 & 99.83 & 99.69  & 99.79 & 99.55 & 99.90 & \textbf{100.00} & 99.59 & \underline{99.97}\\
\colorbox{pc9}{\textcolor{pc9}{A}} Shadows & 195 & 1970 & \underline{99.95} & 98.68 & 98.83 & 93.60 & 71.42 & 96.65 & \textbf{100.00} & \textbf{100.00}  \\
\hline
Overall 			& 5536 & 97940 & 97.66 & 96.81 & 96.93 & 97.55 & 97.82 & \underline{98.63} & 98.47 & \textbf{98.78}\\
Average 	    		&  & & 95.01 & 94.96 & 94.87 & 95.95 & 94.16 & \underline{97.64} & 97.61 & \textbf{98.46}\\
$\kappa$ 		& & & 0.958 & 0.943 & 0.945 & 0.956 & 0.960 & \underline{0.972} & \underline{0.972} & \textbf{0.978}\\
\end{tabular}
\label{tab:results_paviaCenter}
\end{table*} 
    
\section{Conclusion and Outlook}
The paper presented a shapelet-based sparse representation approach with a constructed spatial-spectral dictionary for the classification of hyperspectral image data.
The presented approach differs from previously proposed sparse representation-based classifiers for hyperspectral image data in this way that sophisticated prior knowledge about the spatial nature of an image is exploited by utilizing a constructed, highly adapted patch-specific dictionary.
The experimental results show that our proposed approach outperforms or performed at least equally well in terms of accuracies, when compared to other sparse representation-based classification procedures which use only limited spatial information or state-of-the-art classifiers, which use spatial information.
Moreover, a replacement of image-specific shapelets with Haar wavelets lead to a decrease in accuracy, showing the gain in accuracy when using characteristic spatial patterns rather than synthetic patterns.
It is interesting to underline the stable performance of the proposed \name~ approach, when comparing the results achieved on the different data sets.  Contrary, the other methods show diverse performance in terms of accuracy, when classifying different data sets.
While a method can be applicable for the classification of one data set (\ie, resulting in a high classification accuracy), the same method seems not adequate for the classification of another data set.

Experiments showed the influence of the user-dependent parameters, \ie, the number of shapelets and dictionary elements, on the classification accuracy.
The patchsize can be chosen intuitively according to the spatial homogeneity in the image.
Moreover, the approach seems to be insensitive if a larger patchsize and an adequate number of shapelets is chosen.
Our experimental results allow some guidelines with regard to reliable ranges for the two considered parameters, \ie, according to the results, about $5-100$ shapelets should be used, and a number of three dictionary elements ($W = 3$) appears sufficient. 
These recommendations proved effective in all three experiments.
Therefore the proposed sparse representation-based classification with dictionary construction constitutes a feasible approach and useful modification of the state-of-the-art classifiers utilizing spatial information and kernelized sparse representation based classifiers.

Future work could address the improvement of the computation time for constructing the patch-specific dictionary.
The runtime of our prototype Matlab implementation for the presented datasets using the best parameters settings is about $0.5$h for \textsc{Indian Pines}, about $6$h for \textsc{University of Pavia} and about $25$h for \textsc{Center of Pavia} using a single node (Intel Xeon Westmere X5650 CPU with 12 threads) of the Soroban supercomputer at FU Berlin\footnote{technical details at \url{https://www.zedat.fu-berlin.de/HPC/Soroban}}.
The computational bottleneck is caused by solving an optimization function (see Eq. \ref{eq:energy}) for each dictionary element.  
However, the determination of the dictionary elements is parallelizable.
Speed-up can be achieved by \eg~grouping similar patches and reconstruct them with the same dictionary or using non-fully overlap of image patches for reconstruction, although at the cost of a decrease in accuracy.

\section*{Acknowledgement}
The authors would like to thank D.~Landgrebe and L.~Biehl~(Purdue University, USA) for providing the \textsc{Indian Pines} dataset~(available on: \url{https://engineering.purdue.edu/~biehl/MultiSpec/hyperspectral.html}) and P.~Gamba~(University of Pavia, Italy) for providing the Pavia datasets.
The authors would also like to thank the German Research Foundation (DFG) WA 2728/3-1 for funding and The Fields Institute for Research in Mathematical Sciences, Toronto, Canada.
The research herein was performed in part while Ribana Roscher was visiting the Fields Institute.

\bibliographystyle{IEEEtran}

\begin{IEEEbiography}[{\includegraphics[width=1in,height=1.25in,clip,keepaspectratio]{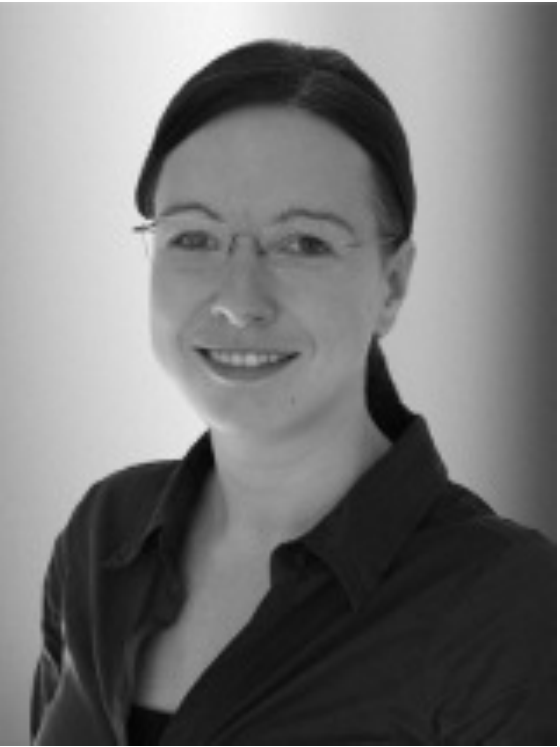}}]{Ribana Roscher}
received the Dipl.-Ing. degree in geodesy from University of Bonn, Germany, in 2008.
She received his Ph.D. in geodesy and geoinformation from the same institution in 2012.
Since 2013 she is a postdoctoral researcher at Freie Universit\"at Berlin, Institute of Geographical Sciences, focussing on methods for land cover classification and sparse representations for high-dimensional sensor data.
In 2015 she was a visiting researcher at The Fields Institute for Research in Mathematical Sciences, University of Toronto, Canada, joining the thematic program on 'Statistical inference, learning, and models for big data'.
Dr. Roscher is reviewer for different international journals, including \textsc{IEEE Transactions on Geoscience and Remote Sensing}, \textsc{IEEE Journal of Selected Topics in Applied Earth Observations and Remote Sensing} and \textsc{IEEE Transactions on Neural Networks and Learning Systems}.

\end{IEEEbiography}

\begin{IEEEbiography}[{\includegraphics[width=1in,height=1.25in,clip,keepaspectratio]{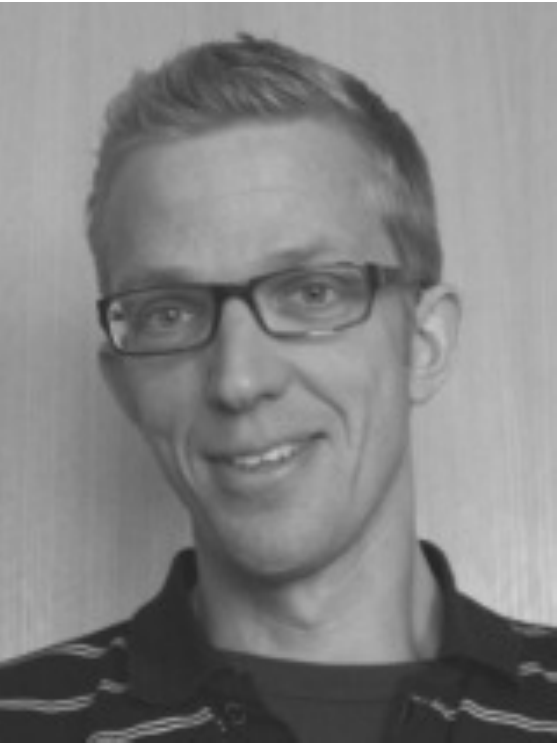}}]{Bj\"orn Waske}
received the degree in applied environmental sciences with a major in remote sensing from Trier University, Germany, in 2002 and the Ph.D. degree in geography from the University of Bonn, Germany, in 2007. 
From 2008 until August 2009, he was a postdoctoral researcher at the Faculty of Electrical and Computer Engineering, University of Iceland, Reyjavik, Iceland; followed by a (junior)professor for remote sensing at the University of Bonn, Germany, until September 2013. 
Since then he is a professor for remote sensing and geoinformatics at the Freie Universität Berlin, Germany. 
His current research activities concentrate on advanced concepts for image classification and data fusion, with a strong focus on monitoring land use cover and land use cover change. 
Dr. Waske is a reviewer for different international journals, including the \textsc{IEEE Transactions on Geoscience and Remote Sensing}, \textsc{IEEE Geoscience and Remote Sensing Letters}, \textsc{IEEE Journal of Selected Topics in Applied Earth Observations and Remote Sensing} and \textsc{Remote Sensing}.
\end{IEEEbiography}

\vfill
\end{document}